\documentclass[10pt,twocolumn,letterpaper]{article}

\usepackage{cvpr}              %

\usepackage{amsmath}
\usepackage{bm}
\usepackage{amssymb}
\usepackage{mathtools}
\usepackage{amsthm}
\usepackage{multirow}
\usepackage{graphicx}
\usepackage{amsmath}
\usepackage{amssymb}
\usepackage{booktabs}
\usepackage{wrapfig}
\usepackage{wrapfig}
\usepackage{float}
\floatstyle{plaintop}
\restylefloat{table}
\usepackage{url} 
\graphicspath{ {figure/} }
\usepackage{amsfonts}

\usepackage{algorithm}
\usepackage{algpseudocode}

\theoremstyle{plain}
\newtheorem{theorem}{Theorem}[section]
\newtheorem{proposition}[theorem]{Proposition}

\theoremstyle{definition}

\theoremstyle{remark}

\usepackage[accsupp]{axessibility} 
\usepackage[pagebackref,breaklinks,colorlinks]{hyperref}

\usepackage[capitalize]{cleveref}
\crefname{section}{Sec.}{Secs.}
\Crefname{section}{Section}{Sections}
\Crefname{table}{Table}{Tables}
\crefname{table}{Tab.}{Tabs.}

\def\cvprPaperID{9569} %
\def\confName{CVPR}
\def\confYear{2023}

\begin{document}

\title{Gradient-based Uncertainty Attribution for Explainable Bayesian Deep Learning}

\author{Hanjing Wang\\
Rensselaer Polytechnic Institute\\
{\tt\small wangh36@rpi.edu}
\and
Dhiraj Joshi\\
IBM Research\\
{\tt\small djoshi@us.ibm.com}
\and
Shiqiang Wang\\
IBM Research\\
{\tt\small wangshiq@us.ibm.com}
\and
Qiang Ji\\
Rensselaer Polytechnic Institute\\
{\tt\small jiq@rpi.edu}
}
\maketitle
\vspace{-15mm}
\begin{abstract}
Predictions made by deep learning models are prone to data perturbations, adversarial attacks, and out-of-distribution inputs. To build a trusted AI system, it is therefore critical to accurately quantify the prediction uncertainties. While current efforts focus on improving uncertainty quantification accuracy and efficiency, there is a need to identify uncertainty sources and take actions to mitigate their effects on predictions. Therefore, we propose to develop explainable and actionable Bayesian deep learning methods to not only perform accurate uncertainty quantification but also explain the uncertainties, identify their sources, and propose strategies to mitigate the uncertainty impacts. Specifically, we introduce a gradient-based uncertainty attribution method to identify the most problematic regions of the input that contribute to the prediction uncertainty. Compared to existing methods, the proposed UA-Backprop has competitive accuracy, relaxed assumptions, and high efficiency. Moreover, we propose an uncertainty mitigation strategy that leverages the attribution results as attention to further improve the model performance. Both qualitative and quantitative evaluations are conducted to demonstrate the effectiveness of our proposed methods.
\end{abstract}
\vspace{-4mm}
\section{Introduction}
\label{sec:intro}
Despite significant progress in many fields, conventional deep learning models cannot effectively quantify their prediction uncertainties, resulting in overconfidence in unknown areas and the inability to detect attacks caused by data perturbations and out-of-distribution inputs. Left unaddressed, this may cause disastrous consequences for safety-critical applications, and lead to untrustworthy AI models.   

The predictive uncertainty can be divided into epistemic uncertainty and aleatoric uncertainty \cite{kendall2017uncertainties}.  %
Epistemic uncertainty reflects the model's lack of knowledge about the input. %
High epistemic uncertainty arises in regions, where there are few or no observations. Aleatoric uncertainty measures the inherent stochasticity in the data. Inputs with high noise are expected to have high aleatoric uncertainty. Conventional deep learning models, such as deterministic classification models that output softmax probabilities, can only estimate the aleatoric uncertainty.

Bayesian deep learning (BDL) offers a principled framework for estimating both aleatoric and epistemic uncertainties.  Unlike the traditional point-estimated models, BDL constructs the posterior distribution of model parameters. By sampling predictions from various models derived from the parameter posterior, BDL avoids overfitting and allows for systematic quantification of predictive uncertainties. However, current BDL methods primarily concentrate on enhancing the accuracy and efficiency of uncertainty quantification, while failing to explicate the precise locations of the input data that cause predictive uncertainties and take suitable measures to reduce the effects of uncertainties on model predictions.

Uncertainty attribution (UA) aims to generate an uncertainty map of the input data to identify the most problematic regions that contribute to the prediction uncertainty. It evaluates the contribution of each pixel to the uncertainty, thereby increasing the transparency and interpretability of BDL models. Previous attribution methods are mainly developed for classification attribution (CA) with deterministic neural networks (NNs) to find the contribution of image pixels to the classification score.  Unlike UA, directly leveraging the gradient-based CA methods for detecting problematic regions is unreliable. While CA explains the model’s classification process, assuming its predictions are confident, UA intends to identify the sources of input imperfections that contribute to the high predictive uncertainties. Moreover, CA methods are often class-discriminative since the classification score depends on the predicted class. As a result, they often fail to explain the inputs which have wrong predictions with large uncertainty \cite{perez2022attribution}. Also shown by Ancona et al. \cite{ancona2017towards}, they are not able to show the troublesome areas of images for complex datasets. Existing CA methods can be categorized into gradient-based methods \cite{simonyan2013deep,zeiler2014visualizing,springenberg2014striving,smilkov2017smoothgrad,sundararajan2017axiomatic,xu2020attribution,kapishnikov2021guided, selvaraju2017grad,srinivas2019full} and perturbation-based methods \cite{ribeiro2016should,petsiuk2018rise,fong2017interpretable,dabkowski2017real, fong2019understanding,yang2021mfpp}. The former directly utilizes the gradient information as input attribution, while the latter modifies the input and observes the corresponding output change. However, perturbation-based methods often require thousands of forward propagations to attribute one image, suffering from high complexity and attribution performance varies for different chosen perturbations. %
Although CA methods are not directly applicable, we will discuss their plain extensions for uncertainty attribution in Sec. \ref{vanilla_adoption}. %

Recently, some methods are specifically proposed for UA. For example, CLUE \cite{antoran2020getting} and its variants \cite{ley2021delta,ley2022diverse} aim at generating a better image with minimal uncertainty by modifying the uncertain input through a generative model, where the attribution map is generated by measuring the difference between the original input and the modified input. Perez et al. \cite{perez2022attribution} further combine CLUE with the path integral for improved pixel-wise attributions. However, these methods are inefficient for real-time applications because they require solving one optimization problem per input for a modified image. Moreover, training generative models is generally hard and can be unreliable for complex tasks. 

We propose a novel gradient-based UA method, named UA-Backprop, to effectively address the limitations of existing methods. The contributions are summarized below. 
\begin{itemize}
    \item UA-Backprop backpropagates the uncertainty score to the pixel-wise attributions, without requiring a pretrained generative model or additional optimization. The uncertainty is fully attributed to satisfy the completeness property, i.e., the uncertainty can be decomposed into the sum of individual pixel attributions. The explanations can be generated efficiently within a single backward pass of the BDL model. 
    \item We introduce an uncertainty mitigation approach that employs the produced uncertainty map as an attention mechanism to enhance the model's performance. We present both qualitative and quantitative evaluations to validate the efficacy of our proposed method. 
\end{itemize}

\section{Preliminaries}
\label{pre}

\subsection{BDL and Uncertainty Quantification}
\label{bdl}
BDL models assume that the neural network parameters $\boldsymbol \theta$ are random variables, with a prior $p(\boldsymbol \theta)$ and a likelihood $p(\mathcal{D}|\boldsymbol \theta)$, where $\mathcal{D}$ represents the training data. We can apply the Bayes' rule to compute the posterior of $\boldsymbol \theta$, i.e., $p(\boldsymbol \theta|\mathcal{D})$ as shown in the following equation:
\begin{equation}
\label{eq:posterior_para}
p(\boldsymbol \theta \mid \mathcal{D}) = \frac{p(\mathcal{D}\mid \boldsymbol \theta)p(\boldsymbol \theta)}{p(\mathcal{D})}.\\
\end{equation}
Computing the posterior analytically is often intractable. %
Therefore, various methods have been proposed for approximately generating parameter samples from the posterior, including MCMC sampling methods \cite{geman1984stochastic, SGHMC_chen2014, hastings1970monte}, variational methods \cite{carvalho2020scalable, louizos_ICML17_MNF, Maddox_NIPS19_SWAG,MacKay1992}, and ensemble-based methods \cite{Lakshminarayanan_NIPS17_ensemble,valdenegro2019deep,huang2017snapshot,wen2020batchensemble,wenzel2020hyperparameter}. The advantages of the BDL models are their capability to quantify aleatoric and epistemic uncertainties. 

Let us denote the input as $\bm x$, the target variable as $\bm y$, and the output target distribution as $p(\bm y|\bm x, \boldsymbol \theta)$ parameterized by $\boldsymbol \theta$, which are the Bayesian parameters such that $\boldsymbol \theta \sim p\left({\boldsymbol \theta} | \mathcal{D}\right)$. In this paper, we will focus on classification tasks. For a given input $\bm x$ and training data $\mathcal{D}$, we estimate the epistemic uncertainty and the aleatoric uncertainty by the mutual information and the expected entropy \cite{depeweg2018decomposition} in: 
\begin{equation}\label{eq:total_entropy}
\begin{split}
  \underbrace{\mathcal{H}\left[p(\bm y | \bm x, \mathcal{D})\right]}_{\text {Total Uncertainty $U_t$}}
  =\!\!\!\!\underbrace{\mathcal{I}\left[\bm y, \boldsymbol \theta | \bm x,\mathcal{D}\right]}_{\text {Epistemic Uncertainty $U_e$}}\!\!\!\!
  +\underbrace{\mathbb{E}_{p\left({\boldsymbol \theta} | \mathcal{D}\right)}\big[\mathcal{H}[p(\bm y | \bm x, \boldsymbol \theta)]\big]}_{\text {Aleatoric Uncertainty $U_a$}}
 \end{split}\vspace{-2mm}
\end{equation}
where $\mathcal{H}$, $\mathcal{I}$, and $\mathbb{E}$ represent the entropy, mutual information, and expectation, respectively. Using Monte Carlo approximation of the posterior, we have
\begin{subequations}
    \label{total uncertainty}
    \begin{align}
        &\mathcal{H}\left[p(\bm y | \bm x, \mathcal{D})\right]=\mathcal{H}\left[\mathbb{E}_{p(\boldsymbol \theta|D)}[p(\bm y| \bm x, \boldsymbol \theta)]\right] \\
        & ~~~~~~~~~~~~~~~~~~~~~~~~~~\approx  \mathcal{H}\left[\frac{1}{S} \sum_{s=1}^S p(\bm y|\bm x,\boldsymbol \theta^s)\right]\\
        &\mathbb{E}_{p\left({\boldsymbol \theta} | \mathcal{D} \right)}\big[\mathcal{H}[p(\bm y | \bm x, \boldsymbol \theta)]\big]\approx \frac{1}{S} \sum_{s=1}^S \mathcal{H}[p(\bm y|\bm x,\boldsymbol \theta^s)]  
    \end{align}
\end{subequations}
where $\boldsymbol \theta^s \sim p\left({\boldsymbol \theta} | \mathcal{D}\right)$ and $S$ is the number of samples.
\begin{figure*}[t]
    \centering    \includegraphics[width=0.92\linewidth]{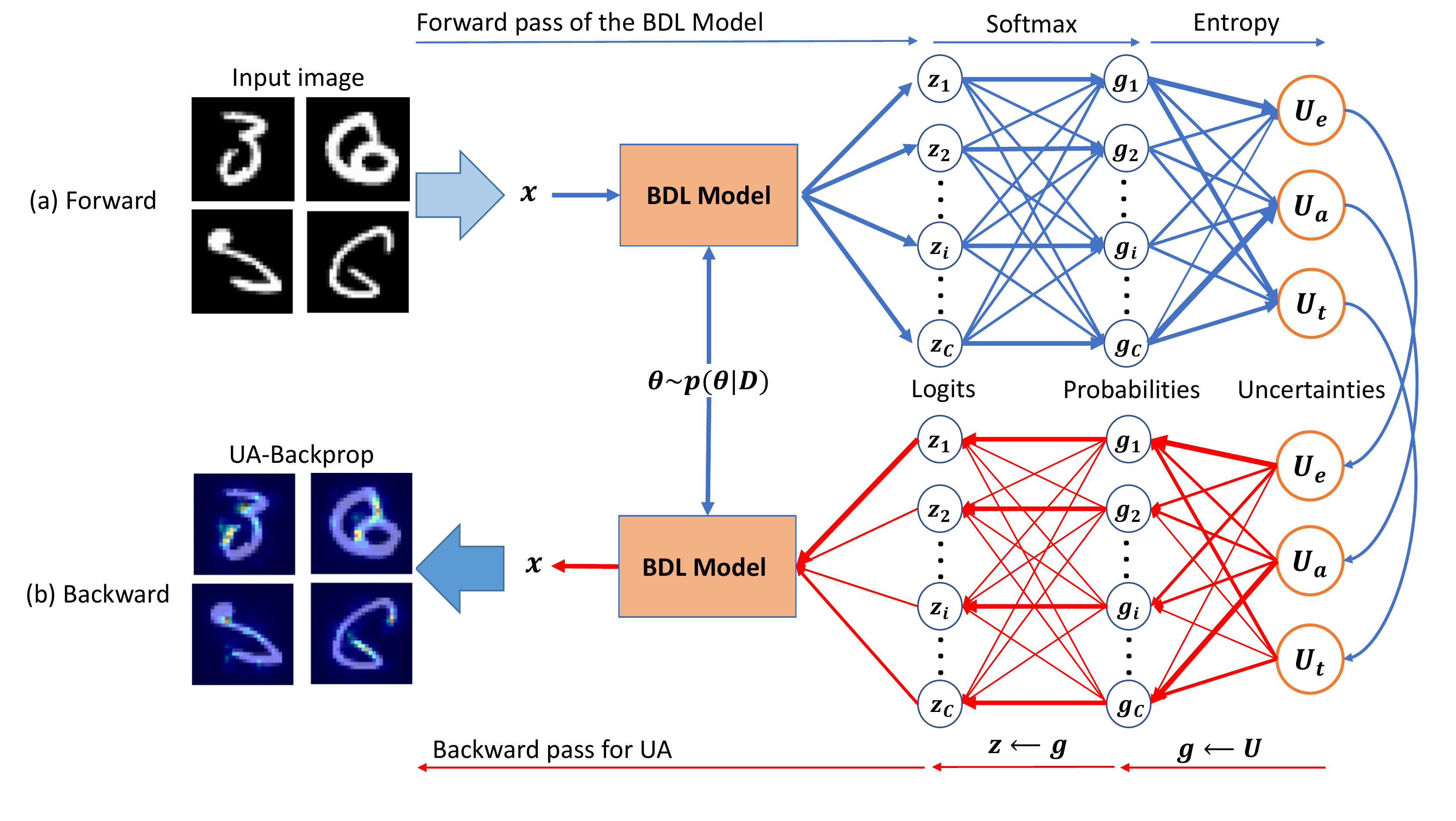}
    \vspace{-4mm}
    \caption{The overall framework of the proposed method. Figure (a) shows the forward propagation of the BDL model for uncertainty quantification. Figure (b) demonstrates the backward process from the uncertainty to the input for attribution analysis, crossing the softmax probabilities, the logits, and the BDL model. The brighter regions indicate higher attributions. \vspace{-4mm}
    \label{fig:overall_framework}}
\end{figure*}

\subsection{Gradient-based Uncertainty Attribution}
\label{vanilla_adoption}
The gradient-based attribution methods can efficiently generate uncertainty maps via backpropagation. While current CA methods mainly utilize the gradients between the model output and input, some of them can be directly extended for UA by using the gradients from the uncertainty to the input. However, raw gradients can be noisy, necessitating the development of various approaches for smoothing gradients, including Integrated Gradient (IG) \cite{sundararajan2017axiomatic} with its variants \cite{xu2020attribution,kapishnikov2021guided}, SmoothGrad \cite{smilkov2017smoothgrad}, Grad-cam \cite{selvaraju2017grad}, and FullGrad \cite{srinivas2019full}. Some methods use layer-wise relevance propagation (LRP) to construct classification attributions. Although the LRP-based methods \cite{bach2015pixel,shrikumar2017learning,montavon2017explaining} can backpropagate the model outputs layer-wisely to the input, there is no direct extension for the uncertainties since we focus on explaining output variations instead of output values. Moreover, they often require specific NN architectures where the entropy and softmax functions for uncertainty estimation will violate their requirements. In this paper, we consider the vanilla extension of SmoothGrad, FullGrad, and IG-based methods as baselines. Please refer to Appendix A and the survey papers \cite{ancona2019gradient,ancona2017towards,nielsen2022robust} for more discussions.  

We contend that the straightforward application of existing attribution methods may not be adequate for conducting UA. Our approach relies on three crucial goals: (1) the uncertainty should be fully attributed with the completeness property satisfied; (2) the pixel-wise attributions should be positive due to data imperfections; (3) the proposed approach should prevent gradient-vanishing issues. Vanilla backpropagation of uncertainty gradients often suffers from the vanishing gradients because of the small magnitude of uncertainty estimates. The resulting visualizations may have ``scatter" attributions, which are incomprehensible. Since vanilla adoption of existing methods for deterministic NNs would always violate some of these goals, 
it is necessary to establish a new gradient-based UA method with competitive accuracy and high efficiency.
\vspace{-1mm}
\section{Uncertainty Attribution with UA-Backprop}

\subsection{Overall Framework}

As shown in Figure \ref{fig:overall_framework}, let $\bm z(\bm x,\boldsymbol\theta) \in \mathcal{R}^C$ denote the output of the neural network with input $\bm x$ parameterized by $\boldsymbol \theta$, which is the probability logit before the softmax layer. The number of classes is represented by $C$. The probability vector $\bm g(\bm x, \boldsymbol \theta)$ is generated from $\bm z(\bm x,\boldsymbol\theta)$ through the softmax function, i.e., $\bm g(\bm x, \boldsymbol \theta) = \mathrm{softmax}(\bm z(\bm x, \boldsymbol \theta))$, where $g_i(\bm x, \boldsymbol \theta) = \frac{\exp(z_i(\bm x, \boldsymbol \theta))}{\sum_{j=1}^C \exp(z_j(\bm x, \boldsymbol \theta))}$. For simplicity, we write $\bm z(\bm x,\boldsymbol\theta)$ as $\bm z$ and $\bm g(\bm x,\boldsymbol \theta)$ as $\bm g$. Since the complex posterior distribution $p(\boldsymbol \theta | \mathcal{D})$ is often intractable, we use a sample-based approximation. We assume that $\{\boldsymbol \theta^s\}_{s=1}^S$ are drawn from $p(\boldsymbol \theta | \mathcal{D})$, leading to samples $\{\bm z^{s}\}_{s=1}^S$ and $\{\bm g^{s}\}_{s=1}^S$. During forward propagation, $\{\bm g^s\}_{s=1}^S$ is used to calculate the epistemic uncertainty $U_e$, aleatoric uncertainty $U_a$, and total uncertainty $U_t$. Let $U$ represent one of the uncertainties in general. For the backpropagation, the uncertainty traverses $U \rightarrow \bm g \rightarrow \bm z \rightarrow \bm x$. The pseudocode for UA-Backprop is provided in Algorithm 1. 
\vspace{-0mm}

\indent  Basically, the contribution of each $g_i$ to $U$, referred to as $U_{g_i}$, is first computed. Since the backward pass of the BDL model contains $S$ paths $\bm g^s \rightarrow \bm z^s \rightarrow \bm x$ for $\boldsymbol \theta^s \sim p(\boldsymbol \theta |\mathcal{D})$, we then obtain the contribution of each $z_i^s$ to $U$, denoted as $U_{z_i^s}$ by exploring all softmax paths $g_j^s \rightarrow z_i^s$ for $j\in [1,\cdots,C]$. Subsequently, $\bm z_i^s \rightarrow \bm x$ is backpropagated. The UA map $M(\bm x)$ is then generated as the pixel-wise contribution to the uncertainty, which aggregates all existing paths $\bm z_i^s \rightarrow \bm x$ with different $s\in [1,\cdots,S]$ and $i\in [1,\cdots,C]$. To fully attribute the uncertainty, the completeness property is enforced on $M(\bm x)$, as shown in Sec. \ref{sec:property}. The backward steps are elaborated in the following sections. 
\begin{algorithm}[ht]
\caption{UA-Backprop + FullGrad}\label{alg:ensemble}
\begin{algorithmic}
\State {\bfseries Input:} A BDL model $\boldsymbol\theta \sim p(\boldsymbol\theta|\mathcal{D})$ with sample approximation $\{\boldsymbol\theta^s\}_{s=1}^S$; Normalization hyperparameter $\tau_1, \tau_2$; The target input $\bm x$ for explanation. 
\State {\bfseries Ouput:} The uncertainty attribution map $M(\bm x)$.
\State {\bfseries Step 1 ($U\rightarrow \bm g$):} Compute the attribution of softmax probabilities $\{U_{g_j}\}_{j=1}^C$ based on Eq.~\eqref{eq:attribution_g}.
\State {\bfseries Step 2 ($\bm g\rightarrow \bm z$):} Based on Eq.~\eqref{eq:compute_u_zi} and Eq.~\eqref{eq:compute_c}, compute the attribution of each logit $U_{z_i}^s$.
\State {\bfseries Step 3 ($\bm z\rightarrow \bm x$):} Generate the uncertainty attribution map with the aggregation from all paths $\bm z_i^s \rightarrow \bm x$ based on Eq.~\eqref{eq:attribution_x} and Eq.~\eqref{localization}.
\end{algorithmic} 
\end{algorithm}
\subsection{Attribution of Softmax Probabilities}
In this section, we calculate the attribution of $\bm g$ to uncertainty $U$. For any $i$, we denote the contribution of $g_i$ to $U_e$, $U_a$, and $U_t$ as $U_{e,g_i}$, $U_{a,g_i}$, and $U_{t,g_i}$, respectively. In general, we denote $U_{g_i}$ as the attribution of $g_i$ to $U$. By utilizing Eq.~\eqref{total uncertainty}, we can express $U_e$, $U_a$, and $U_t$ in terms of $\{\bm g^{s}\}_{s=1}^S$, and subsequently decompose them into the sum of individual attributions, as shown in the following equation: \vspace{-1em}
\begin{subequations}
    \label{eq:attribution_g}
    \begin{align}
        U_{t,g_i} &= -\left(\frac{1}{S} \sum_{s=1}^S g_i^s\right) \log\left(\frac{1}{S} \sum_{s=1}^S g_i^s\right)   \\
        U_{a,g_i} &=  \frac{1}{S} \sum_{s=1}^S -g_i^s \log g_i^s \\
        U_{e,g_i} &= U_{t,g_i} - U_{a,g_i} ,
    \end{align}
\end{subequations}
In general, we can observe that $U_{e,g_i}$, $U_{a,g_i}$, $U_{t,g_i}$ only depend on $g_i$ and are independent of other elements of $\bm g$. Moreover, the uncertainties are completely attributed to the softmax probability layer, i.e., $U_t = \sum_{i=1}^C U_{t,g_i}$, $U_a = \sum_{i=1}^C U_{a,g_i}$, $U_e = \sum_{i=1}^C U_{e,g_i}$. 
When backpropagating the path $\bm g^s \rightarrow \bm z^s$ to get the attribution of logits, $U_{g_i}$ is shared across samples $\{g_i^s\}_{s=1}^S$.  

\subsection{Attribution of Logits}
In this section, we aim to derive $U_{e,z_i^s}$, $U_{a,z_i^s}$, and $U_{t,z_i^s}$ as the contribution of $z_i^s$ to $U_e$, $U_a$, and $U_t$ by investigating the path from $\bm g^s$ to $\bm z^s$. We introduce $c_{g_j^s \rightarrow z_i^s} \in (0,1)$ as the coefficient that represents the proportion of the uncertainty attribution that $z_i^s$ receives from $g_j^s$. Through collecting all the messages from $\{g_j^s\}_{j=1}^C$, the contribution of $z_i^s$ to $U$, donated as $U_{z_i^s}$, is a weighted combination of the attributions received from the previous layer: \vspace{-1mm}
\begin{equation}
\label{eq:compute_u_zi}
    U_{z_i^s} = \sum_{j=1}^C c_{g_j^s \rightarrow z_i^s} U_{g_j}.
\end{equation}
\indent To satisfy the completeness property, it is expected that $U_{g_j}$ is fully propagated into the logit layer as shown in the following equation: 
\begin{equation}
\label{eq:completeness_g_to_z}
    U_{g_j}  = \sum_{i=1}^C c_{g_j^s \rightarrow z_i^s} U_{g_j},
\end{equation} 
which is a commonly held assumption in many message-passing mechanisms. Eq.~\eqref{eq:completeness_g_to_z} indicates that $\sum_{i=1}^C c_{g_j^s \rightarrow z_i^s} = 1$. In this paper, we apply the softmax gradients to determine $c_{g_j^s \rightarrow z_i^s}$ for the backward step from $\bm g^s$ to $\bm z^s$. Specifically, the gradient of $g_j^s$ to $z_i^s$ is as follows: 
\begin{equation}
\label{eq:softmax_grad}
    \frac{\partial g_j^s}{\partial z_i^s} =   
    \begin{cases}
    g_j^s(1-g_j^s)      & \quad \text{if } i=j\\
    -g_i^sg_j^s  & \quad \text{if } i\neq j
  \end{cases}.
\end{equation}
Since $\sum_{k=1}^C g_k^s =1$ due to the definition of softmax function, it is notable that $|\frac{\partial g_i^s}{\partial z_i^s}| > | \frac{\partial g_j^s}{\partial z_i^s}|$ for $i \neq j$, signifying that $g_i^s$ is the primary source of the attribution for $z_i^s$. We normalize the gradients to the obtain the coefficients using $\phi(\cdot)$, with the aim of circumventing extremely small coefficients and thus addressing the gradient-vanishing problem. In this study, $\phi(\cdot)$ is a softmax function with temperature $\tau_1$, i.e.,  
\begin{equation}
\label{eq:compute_c}
\begin{split}
    c_{g_j^s \rightarrow z_i^s} &= \phi_i\left(\left\{\frac{\partial g_j^s}{\partial z_k^s}\right\}_{k=1}^C, \tau_1\right) \\
    &= \frac{\exp\left(\frac{\partial g_j^s}{\partial z_i^s}/(g_j^s\cdot \tau_1)\right)}{\sum_{k=1}^C \exp\left(\frac{\partial g_j^s}{\partial z_k^s}/(g_j^s \cdot \tau_1)\right)},
\end{split}
\end{equation}
where $g_j^s \cdot \tau_1$ is employed for avoiding uniform or extremely small coefficients. It is expected that $g_i^s$ provides the major contribution to $z_i^s$ since the denominator of the softmax function in $\bm z^s \rightarrow \bm g^s$ serves only as a normalization term.

\subsection{Attribution of Input}
Given the uncertainty attribution $\{U_{z_i^s}\}_{i=1}^C$, associated with $\{z_i^s\}_{i=1}^C$, the attribution map in the input space is generated by backpropagating through $\bm z^s \rightarrow \bm x$. Since each $z_i^s$ may represent different regions of the input, we individually find the corresponding regions of $\bm x$ that contribute to each $z_i^s$, denoted by $M_i^s(\bm x)$. Finally, the uncertainty attribution map $M(\bm x)$ is derived by a linear combination of $M_i^s(\bm x)$ and $U_{z_i^s}$, i.e.,
\begin{equation} 
    \label{eq:attribution_x}
    M(\bm x) = \frac{1}{S} \sum_{s=1}^S \sum_{i=1}^C U_{z_i}^s M_i^s(\bm x). 
\end{equation}
$M(\bm x)$ indicates the pixel-wise attributions of $U$, which is a two-dimensional matrix that has the same height and width as $\bm x$. It is worth noting that during exploring the possible paths for aggregation, the noisy gradients may be smoothed. We notice that some existing gradient-based methods can be used for exploring the path $\bm z^s \rightarrow \bm x$. For example, the magnitude of the raw gradient can be employed such that $M_i^s(\bm x) = |\frac{\partial z_i^s}{\partial \bm x}|$. Especially, more advanced gradient-based methods such as SmoothGrad \cite{smilkov2017smoothgrad}, Grad-cam \cite{selvaraju2017grad}, and FullGrad \cite{srinivas2019full} can be applied. Intuitively, our proposed method can be a general framework. For the FullGrad method as an example, it aggregates both the gradient of $z_i^s$ with respect to input ($\frac{\partial z_i^s}{\partial \bm x}$) and the gradient of $z_i^s$ with respect to the bias variable $\bm b_{l}^s$ in each convolutional or fully-connected layer $l$ (i.e., $\frac{\partial z_i^s}{\partial \bm b_l^s}$) to create $M_i^s(\bm x)$, i.e.,
\begin{equation}
\label{localization}
    M_{i}^s(\bm x) = \psi\left(\left|\frac{\partial z_i^s}{\partial \bm x} \odot \bm x \right| + \sum_{l}  \left|\frac{\partial z_i^s}{\partial \bm b_l^s}\odot \bm b_{l}^s\right|, \tau_2\right),
\end{equation}
where $\odot$ is the element-wise product and $|\cdot|$ returns the absolute value. Since different methods will have different scales of $M_i(\bm x)$, we apply a post-processing function $\psi$ for normalizing and rescaling the gradients. The function $\psi$ first averages over the channels of $\left|\frac{\partial z_i^s}{\partial \bm x} \odot \bm x \right| + \sum_{l}  \left|\frac{\partial z_i^s}{\partial \bm b_l^s}\odot \bm b_{l}^s\right|$ and then applies an element-wise softmax function with temperature $\tau_2$. As a general framework, we can leverage the current development of gradient-based attribution methods for deterministic NNs to smooth the gradients and avoid the gradient-vanishing issue. 

\subsection{Special Properties}
\label{sec:property}
Our proposed method satisfies the completeness property, shown in the following equation:
\begin{equation}
\label{eq:completeness}
    U = \sum_{i=1}^C U_{g_i} = \sum_{i=1}^C U_{z_i^s} = \sum_{(u,v)} M(\bm x)[u,v],\vspace{-1mm}
\end{equation}
where $(u,v)$ is the index for the entries of $M(\bm x)$. The proof can be found in Appendix A. Our method can also be used with various sensitivity methods for $\bm z \rightarrow \bm x$ to satisfy different properties such as implementation invariance and linearity, which are detailed in Appendix A.

\section{Uncertainty Mitigation}
\label{mitigation}
\begin{figure}[h]
    \centering
    \includegraphics[width=1.0\linewidth]{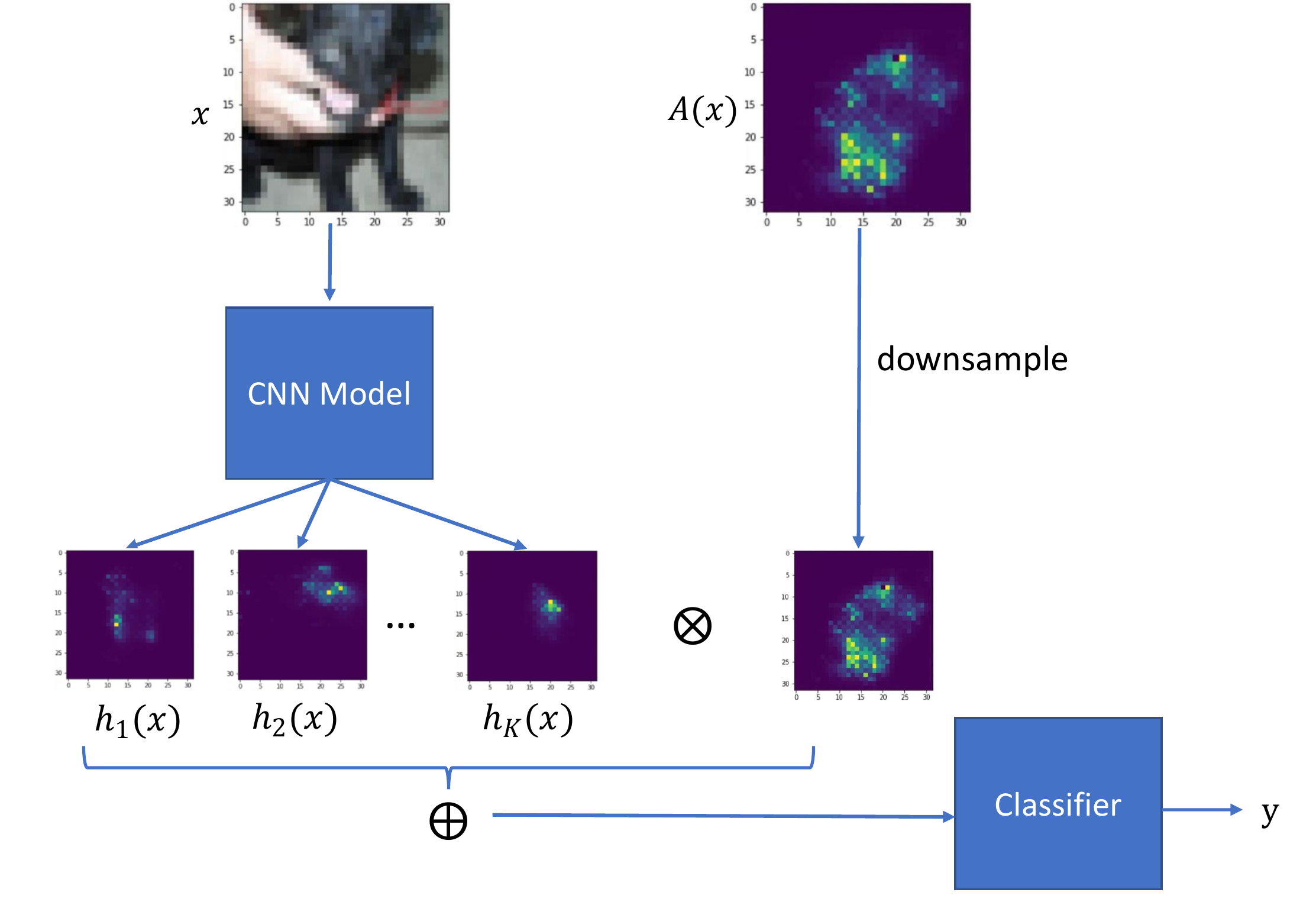}
    \caption{The uncertainty mitigation with attention mechanism.  
    \label{fig:attention_process}} \vspace{-3mm}
\end{figure}

Leveraging the insights gained from uncertainty attribution, uncertainty mitigation is to develop an uncertainty-driven mitigation strategy to enhance model performance. In particular, the uncertainty attribution map $M(\bm x)$ can be utilized as an attention mechanism by multiplying the inputs or features with $1-M(\bm x)$. This can help filter out problematic input information and improve prediction robustness. However, this approach also assigns high weights to unessential background pixels, which is undesirable. To address this issue, the attention weight $A(\bm x)$ is defined by the element-wise product of $(1-M(\bm x))$ and $M(\bm x)$ in order to strengthen more informative areas, as shown as follows:
\begin{equation}
\label{eq:a(x)}
    A(\bm x) = (1-M(\bm x)) \odot M(\bm x).
\end{equation}
It is important to note that the attention mechanism can be implemented either in the input space or in the latent space. In this study, we apply $A(\bm x)$ in the latent space, while conducting ablation studies for the input-space attentions in Sec.~\ref{sec:ablation}. Let $\{\bm h_k(\bm x)\}_{k=1}^K$ with size $K$ be the 2D feature maps generated by the last convolutional layer. We downsample $A(\bm x)$ to match the dimensions of $\bm h_k(\bm x)$ and utilize $\{(1+\alpha A(\bm x)) \odot \bm h_k(\bm x)\}_{k=1}^K$ as inputs to the classifier, where $\alpha$ is a hyperparameter that can be tuned. Through retraining using the masked feature maps, the model gains improved accuracy and robustness by ignoring the unimportant background information and the fallacious regions. The complete process is illustrated in Figure \ref{fig:attention_process}. 

\begin{figure*}[ht]
    \centering
    \includegraphics[width=0.98\linewidth]{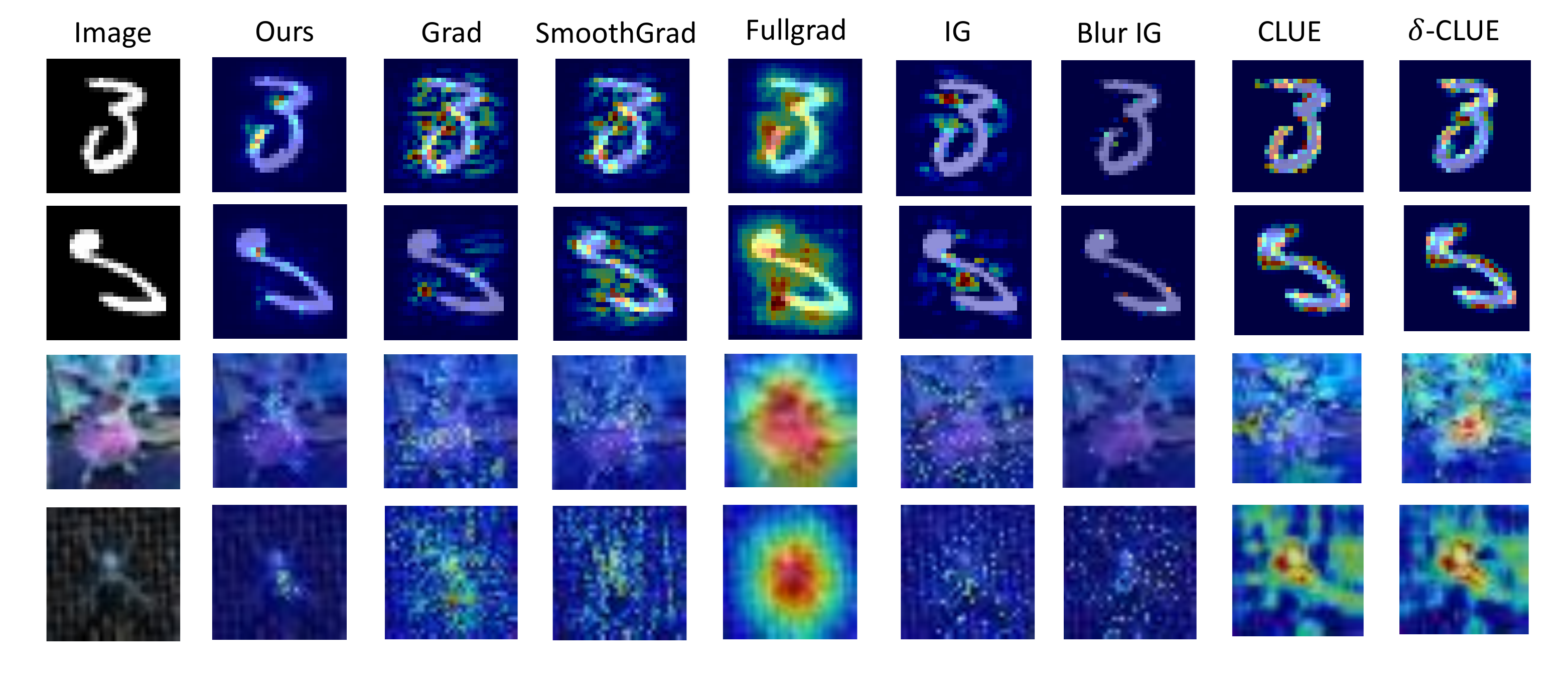}\vspace{-5mm}
    \caption{Examples of the epistemic uncertainty attribution maps for various methods on different datasets. Brighter areas indicate essential regions that contribute most to the uncertainty. More examples can be found in Appendix E.}
    \label{fig:qualitative}\vspace{-5mm}
\end{figure*}

\section{Experiments}
\noindent \textbf{Dataset.} We evaluate the proposed method on the benchmark image classification datasets including MNIST \cite{deng2012mnist}, SVHN \cite{netzer2011reading}, CIFAR-10 (C10) \cite{krizhevsky2014cifar}, and CIFAR-100 (C100)~\cite{krizhevsky2009learning}. 

\noindent \textbf{BDL Model.} In our experiments, we use the deep ensemble method \cite{Lakshminarayanan_NIPS17_ensemble} for uncertainty quantification, which trains an ensemble of deep neural networks from random initializations. It demonstrates great success in predictive uncertainty calibration and outperforms various approximate Bayesian neural networks \cite{Lakshminarayanan_NIPS17_ensemble}. %

\noindent \textbf{Implementation Details.} We use standard CNNs for MNIST/SVHN and Resnet18 for C10/C100. The experiment settings, implementation details, and hyperparameters are provided in Appendix B. 

\noindent \textbf{Baselines.} We compare our proposed method (UA-Backprop + FullGrad) with various baselines on gradient-based uncertainty attribution. The baselines include the vanilla extension of Grad \cite{simonyan2013deep}, SmoothGrad \cite{smilkov2017smoothgrad}, FullGrad \cite{srinivas2019full}, IG \cite{sundararajan2017axiomatic}, and Blur IG \cite{xu2020attribution} for UA. Although CLUE-variants require a generative model and have low efficiency, we include CLUE \cite{antoran2020getting} and $\delta$-CLUE \cite{ley2021delta} for comparison. %

\noindent \textbf{Evaluation Tasks.} In Sec. 5.1, we qualitatively evaluate the UA performance. In Sec. 5.2, we provide the quantitative evaluations including the blurring test, and the attention-based uncertainty mitigation. Various supplementary studies are provided in Appendices C and D.

\subsection{Qualitative Evaluation}
Figure \ref{fig:qualitative} exhibits various examples of attribution maps generated using different techniques. Our analysis reveals that vanilla adoption of CA methods may not be sufficient to generate clear and meaningful visualizations. For instance, as illustrated in Figure \ref{fig:qualitative}, we may expect the digit ``3" to have a shorter tail, the digit ``9" to have a hollow circle with a straight vertical line, and the face of the dog and the small dark body of the spider to be accurately depicted. However, methods such as Grad and Smoothgrad produce ambiguous explanations due to noisy gradients, while FullGrad employs intermediate hidden layers' gradients to identify problematic regions but often lacks detailed information and overemphasizes large central regions. Furthermore, CLUE-based methods tend to identify multiple boundary regions as problematic. They may also fail to provide a comprehensive explanation for complex datasets, where generative models may face significant difficulties in modifying the input to produce an image with lower uncertainty. Finally, CLUE-based methods, Grad, SmoothGrad, and FullGrad fail to fully attribute the uncertainty through the decomposition of pixel-wise contributions. While IG-based methods satisfy the completeness property if the starting image has zero uncertainty, they often produce scattered attributions with minimal regional illustration, posing difficulties in interpretation. 
\begin{figure}[h]
    \centering
    \includegraphics[width=0.75\linewidth]{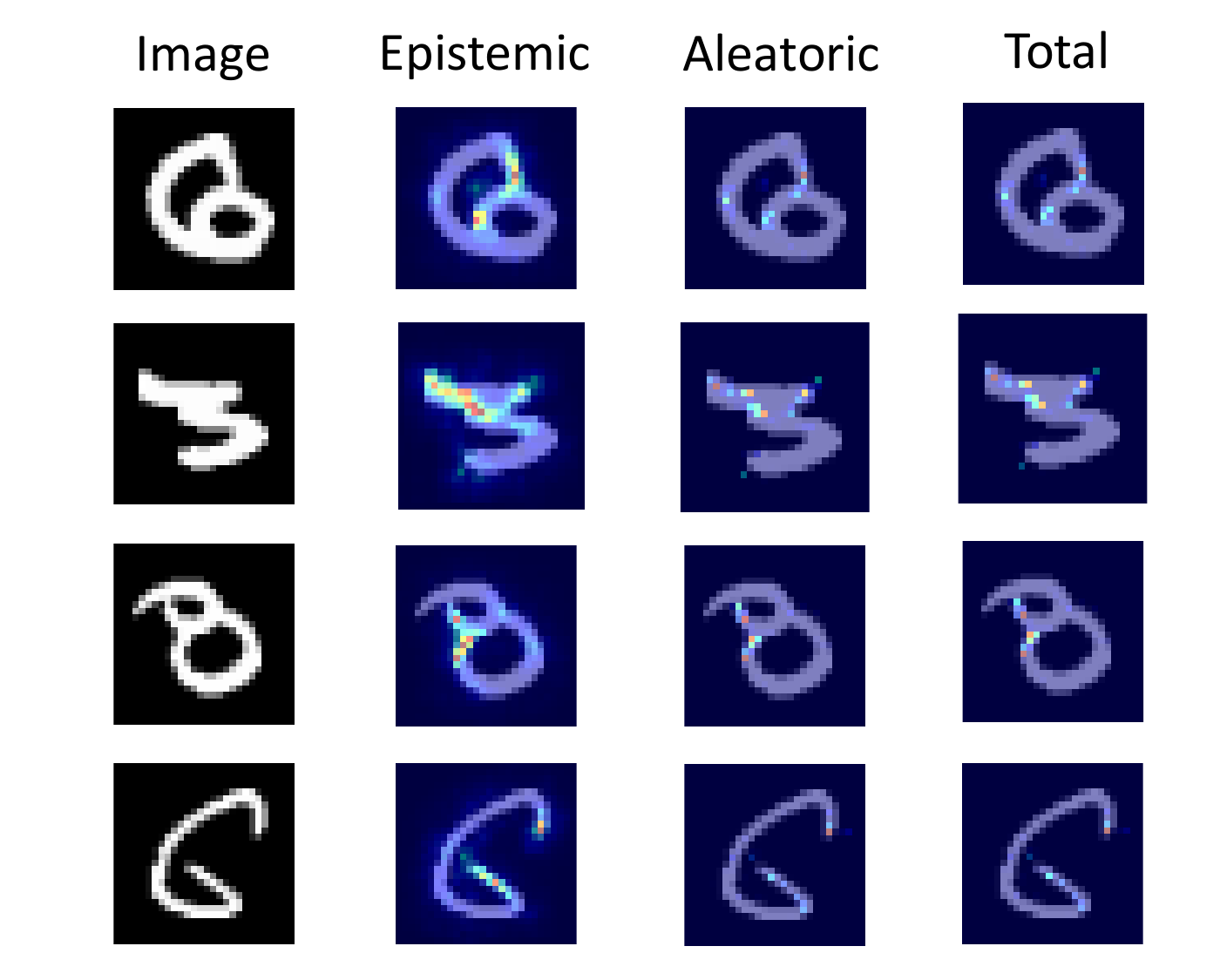}
    \caption{Epistemic, aleatoric, and total uncertainty attribution maps for our proposed method on MNIST dataset.
    \label{fig:different_u}} \vspace{-5mm}
\end{figure}

\begin{table*}[t]
\fontsize{9.25}{12}\selectfont
	\caption{Attribution performance in terms of MURR and AUC-URR. We evaluate on four different datasets and blur the image with a maximum of 2\% or 5\% pixels with the highest contribution to the epistemic uncertainty. The bold values indicate the best performance.}
	\label{tab:blur_test}
	\centering
\begin{tabular}{|l|ccccccccc|l|}
\hline
	\multirow{3}{*}{Method} & \multicolumn{9}{|c|}{Maximum Uncertainty Reduction Rate (MURR) $\uparrow$ }\\ 
 	\cline{2-10}
	& \multicolumn{2}{c}{MNIST} & \multicolumn{2}{c}{C10} & \multicolumn{2}{c}{C100}& \multicolumn{2}{c}{SVHN}  & \multirow{1}{*}{Avg. Performance} \\
 &\%2 &\%5 &\%2 &\%5 &\%2 &\%5 &\%2 &\%5  &\%2 + \%5 \\
\hline  
\multirow{1}{*}{Ours}  &0.648&0.850&0.629&0.848&\textbf{0.195}&0.302&0.625&0.758&\textbf{0.607}\\
\multirow{1}{*}{Grad}  &0.506&0.741&0.578&0.798&0.165&0.276&0.555&0.705&0.541\\
\multirow{1}{*}{SmoothGrad}  &0.601&0.779&0.566&0.800&0.154&0.255&0.575&0.735&0.558\\
\multirow{1}{*}{FullGrad}  &\textbf{0.691}&0.869&0.555&0.772&0.156&0.274&0.565&0.709&0.574\\
\multirow{1}{*}{IG}  &0.434&0.725&0.632&0.827&0.159&0.270&0.649&0.773&0.559\\
\multirow{1}{*}{Blur IG} &0.305&0.515&\textbf{0.693}&\textbf{0.971}&0.184&\textbf{0.318}&\textbf{0.762}&\textbf{0.896}&0.581\\
\multirow{1}{*}{CLUE}  &0.614&0.874&0.291&0.628&0.074&0.148&0.171&0.352&0.394\\
\multirow{1}{*}{$\delta$-CLUE} &0.625&\textbf{0.901}&0.415&0.577&0.073&0.150&0.146&0.295&0.398\\
\hline
\end{tabular}\vspace{1mm}
\begin{tabular}{|l|ccccccccc|l|}
\hline
	\multirow{3}{*}{Method} & \multicolumn{9}{|c|}{Area under the Uncertainty Reduction Curve (AUC-URR) $\downarrow$ }\\ 
 	\cline{2-10}
	& \multicolumn{2}{c}{MNIST} & \multicolumn{2}{c}{C10} & \multicolumn{2}{c}{C100}& \multicolumn{2}{c}{SVHN} & \multirow{1}{*}{Avg. Performance} \\
 &\%2 &\%5 &\%2 &\%5 &\%2 &\%5 &\%2 &\%5  &\%2 + \%5 \\
\hline
\multirow{1}{*}{Ours}  &0.667&0.445&0.664&0.484&\textbf{0.901}&\textbf{0.821}&0.526&0.407&\textbf{0.614}\\
\multirow{1}{*}{Grad}  &0.709&0.534&0.701&0.538&0.912&0.843&0.613&0.448&0.662\\
\multirow{1}{*}{SmoothGrad}  &0.675&0.461&0.730&0.551&0.919&0.860&0.584&0.424&0.651\\
\multirow{1}{*}{FullGrad}  &\textbf{0.603}&0.429&0.696&0.543&0.924&0.859&0.596&0.455&0.638\\
\multirow{1}{*}{Blur IG}  &0.816&0.667&\textbf{0.638}&0.466&0.914&0.851&0.541&0.402&0.662\\
\multirow{1}{*}{IG} &0.752&0.529&0.731&\textbf{0.444}&0.905&0.824&\textbf{0.523}&\textbf{0.298}&0.626\\
\multirow{1}{*}{CLUE}  &0.709&0.397&0.861&0.624&0.966&0.926&0.919&0.815&0.777\\
\multirow{1}{*}{$\delta$-CLUE}&0.665&\textbf{0.395}&0.793&0.710&0.968&0.924&0.932&0.848&0.779\\
\hline
\end{tabular}\vspace{-3mm}
\end{table*}
\indent Figure \ref{fig:different_u} presents various examples of UA maps that depict different types of uncertainties. It is a well-known fact that epistemic uncertainty inversely relates to training data density. %
Hence, the epistemic uncertainty maps indicate the areas that deviate from the distribution of training data. In some cases, inserting or blurring pixels will help to reduce uncertainty for performance improvement. The aleatoric uncertainty maps quantify the contribution of input noise to prediction uncertainty, which tends to assign high attributions to object boundaries. As displayed in Figure \ref{fig:different_u}, the total uncertainty maps are quite similar to the aleatoric uncertainty maps. That is because the aleatoric uncertainty quantified in Eq.~\eqref{eq:total_entropy} is often much larger than the epistemic uncertainty, which dominates the total uncertainty.
\subsection{Quantitative Evaluation}
\subsubsection{Blurring Test}
\indent \indent Following \cite{perez2022attribution}, we evaluate the proposed method through the blurring test. If the most problematic regions are blurred for a highly uncertain image, we expect a significant uncertainty reduction due to the removal of misleading information. The blurring can be conducted via a Gaussian filter with mean 0 and standard derivation $\sigma$. We iteratively blur the pixels based on their contributions to the uncertainty, where we evaluate the corresponding uncertainty reduction curve to demonstrate the effectiveness of our proposed method. Some examples are shown in Figure \ref{fig:blur} and the detailed experiment setting is shown in Appendix B.

The evaluation for the blurring test is conducted on the epistemic uncertainty map  since the aleatoric uncertainty captures the input noise and is likely to increase when blurring the image. %
\begin{figure}[b]
    \centering
    \includegraphics[width=1.0\linewidth]{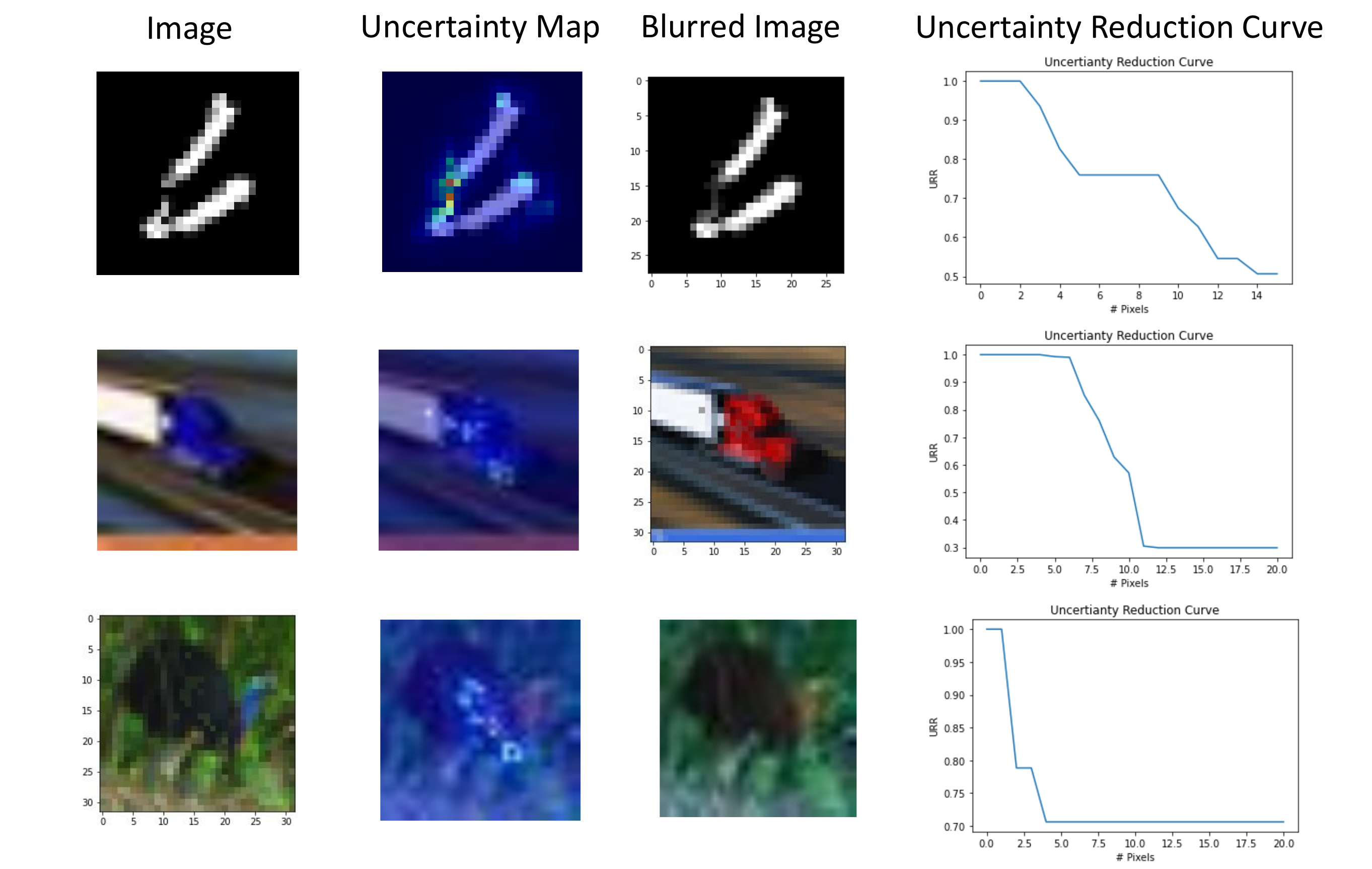}\vspace{-2mm}
    \caption{ Examples of the blurring test for UA-Backprop. 
    \label{fig:blur}}
\end{figure}
Denote $v_1, v_2, \cdots, v_T$ as the pixels that contribute most to the epistemic uncertainty, following the decreasing order. We iteratively blur up to $t$ pixels, i.e., $\bm v_{1:t}$, and denote the resulting blurred image as $\bm x_{t}$. The uncertainty reduction rate (URR) shown in Eq.~\eqref{eq:URR} quantifies the extent of achieved uncertainty reduction for blurring up to $t$ problematic pixels:
\begin{equation}
\label{eq:URR}
        \mathrm{URR}(t) =  \frac{1}{|\mathcal{X}|} \sum_{\bm x\in \mathcal{X}} \max_{i\leq t} 1- \frac{U(\bm x_{i})}{U(\bm x)}.
\end{equation}
For URR, we aggregate the results for various sampled images $\bm x\in \mathcal{X}$. The URR curve, obtained by plotting the decreasing normalized values of $\{\mathrm{URR}(t)\}_{t=1}^T$, is a key performance metric. We report two evaluation metrics, namely, the maximum uncertainty reduction rate (MURR), i.e., $\max_{t=1:T} \mathrm{URR}(t)$, and the area under the URR curve (AUC-URR). Larger MURR and smaller AUC-URR values indicate superior performance of the UA method. Since the blurring may lead some images to be out-of-distribution, we report median values instead. 

\begin{table*}[ht]
\fontsize{9.25}{12}\selectfont
	\caption{Acc (\%) $\uparrow$ and NLL $\downarrow$ for uncertainty mitigation evaluation. The results are aggregated over 5 independent runs. }
	\label{tab:attention_test}
	\centering
\begin{tabular}{|l|cccccccccc|l|}
\hline
	\multirow{2}{*}{Method} 
	& \multicolumn{2}{c}{MNIST} & \multicolumn{2}{c}{C10} & \multicolumn{2}{c}{C100}& \multicolumn{2}{c}{SVHN} & \multicolumn{2}{c|}{Avg. Performance} \\
 &ACC &NLL &ACC &NLL &ACC &NLL &ACC &NLL &ACC &NLL \\
\hline  
\multirow{1}{*}{Ours}  &91.95 &\textbf{0.287} &\textbf{36.48} &\textbf{1.768} &12.12 &4.326 &\textbf{65.13} &\textbf{1.489} &\textbf{51.42}&\textbf{1.968}\\
\multirow{1}{*}{Grad}  &91.35 &0.302 &31.60 &1.938 &12.13& 4.422&63.74&1.578&49.71&2.060\\
\multirow{1}{*}{SmoothGrad}  &90.68 & 0.324 &32.05 &1.942 &\textbf{12.57} &4.508 &62.35 &1.628&49.41&2.100\\
\multirow{1}{*}{FullGrad}  &91.39 &0.300 &32.85 &1.920 &12.06 &4.574 &62.38 &1.568 &49.67&2.091\\
\multirow{1}{*}{IG}  &\textbf{91.98}&0.350&34.43&1.829&11.89&\textbf{4.265}&64.31&1.511&50.65&1.989\\
\multirow{1}{*}{Blur IG}  &91.57 &0.288 &32.20 &1.935 &12.34&4.630 &65.04 &1.526 &50.29 &2.095\\
\multirow{1}{*}{CLUE}  &91.64&0.348&33.34&1.846&12.15&4.299&60.01&1.572&49.29&2.016\\
\multirow{1}{*}{$\delta$-CLUE}  &91.76&0.350&35.02&1.809&12.22&4.362&62.71&1.612&50.43&2.033\\
\multirow{1}{*}{No attention}  &90.78 &0.358 &31.62&1.921&12.02&4.536&60.64&1.569&48.77&2.096 \\
\hline
\end{tabular}\vspace{-3mm}
\end{table*}

As shown in Table \ref{tab:blur_test}, our proposed method achieves the best average performance and ranks among the top three in all datasets. In particular, it consistently outperforms Grad, SmoothGrad, FullGrad, and IG. While Blur IG shows promising performance on certain datasets such as C10 and SVHN, it requires a larger number of blurred pixels to achieve improvements and has no advantages to identify the highest problematic regions. Generative-model-based methods, such as CLUE and $\delta$-CLUE, perform well on MNIST but face difficulties in attributing complex images. Additionally, SmoothGrad, Blur IG, and IG require multiple backward passes to attribute one input, while CLUE and $\delta$-CLUE also require a specific optimization process per image, which makes them less efficient. Overall, our proposed method demonstrates superior performance and stands out as the optimal approach for UA in the blurring test.

\subsubsection{Uncertainty Mitigation Evaluation}
\begin{figure}[h]
    \centering
    \includegraphics[width=0.8\linewidth]{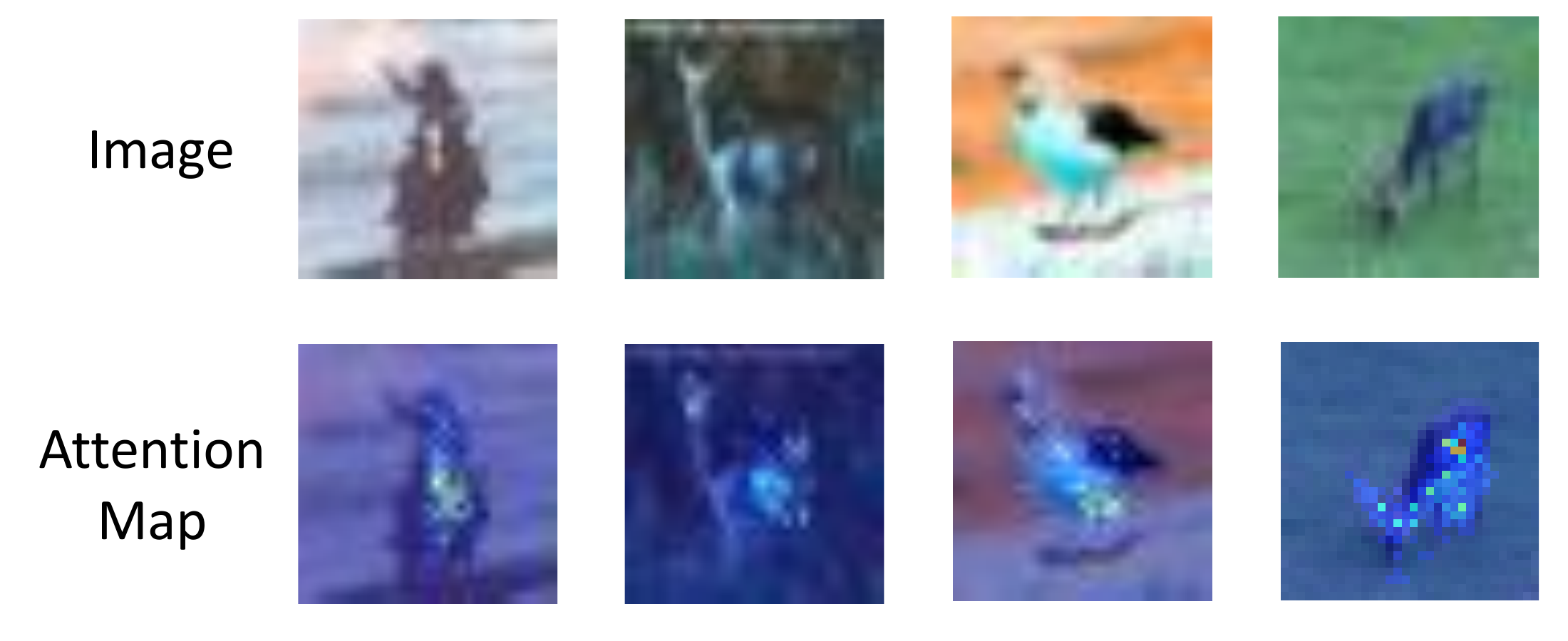}\vspace{-3mm}
    \caption{Examples of attention maps for UA-Backprop.  
    \label{fig:attention}}\vspace{-5mm}
\end{figure}

\indent \indent Building on the methodology in Sec.~\ref{mitigation}, we adopt pre-generated attribution maps as attention mechanisms to enhance model performance. The formulation of attention, denoted by $A(\bm x)$, is presented in Eq.~\eqref{eq:a(x)}, and is exemplified in Figure \ref{fig:attention}. To ensure consistency in scale across different methods, the attribution map $M(\bm x)$ is normalized using the element-wise softmax function before being used in Eq.~\eqref{eq:a(x)}.%

The experimental focus is on training with limited data due to the time-consuming process of generating attribution maps for large datasets, particularly for methods such as Blur IG, SmoothGrad, and CLUE. To this end, we randomly select 500, 1000, 2000, and 4000 images from MNIST, C10, SVHN, and C100, respectively. The selected samples are trained with pre-generated attention maps and evaluated on the original testing data. %
The evaluation metrics used are accuracy (ACC) and negative log-likelihood (NLL). The experimental setup is detailed in Appendix B.

Table \ref{tab:attention_test} presents the results obtained for uncertainty mitigation. The method ``no attention" refers to plain training without attention incorporated. Our method demonstrates a 6\% improvement in ACC compared to vanilla training, suggesting a promising potential for utilizing attribution maps for further model refinement. Our method consistently outperforms other attribution methods in terms of averaged ACC and NLL. We notice that more significant improvement in NLL often occurs for smaller datasets, whereas C100 is challenging to fit with limited samples, and the performance will be more influenced by stochastic training.

\subsubsection{Ablation Studies and Further Analysis} \label{sec:ablation}

\indent \indent To have a comprehensive evaluation, we conduct the anomaly detection experiment in Appendix C, which compares the predicted problematic regions with the known ground truth. Ablation studies such as efficiency analysis, attribution performances under different experiment settings, and hyperparameter sensitivity analysis are provided in Appendix D.

\section{Conclusion}
This research aims at developing explainable uncertainty quantification methods for BDL. It will significantly advance the current state of deep learning, allowing it to accurately characterize its uncertainty and improve its performance, facilitating the development of safe, reliable, and trustworthy AI systems. Our proposed method is designed to attribute the uncertainty to the contributions of individual pixels within a single backward pass, resulting in competitive accuracy, relaxed assumptions, and high efficiency. The results of both qualitative and quantitative evaluations suggest that our proposed method has a high potential for producing dependable and comprehensible visualizations and establishing mitigation strategies to reduce uncertainty and improve model performance.
\\

\noindent \textbf{Acknowledgement:} This work is supported in part by DARPA grant  FA8750-17-2-0132 and by the Rensselaer-IBM AI Research Collaboration (\url{http://airc.rpi.edu}), part of the IBM AI Horizons Network.

% {\small
% \bibliographystyle{ieee_fullname}
% \bibliography{main}
% }

\clearpage

\appendix

\section{Properties of UA Methods}
In this section, we present an overview of the key characteristics of uncertainty attribution methods, which are extended from the attribution methods for deterministic NNs. We also provide a brief introduction to various existing gradient-based attribution methods for deterministic NNs, along with their vanilla extensions. Lastly, we provide a theoretical demonstration of the essential properties of our proposed method.
\subsection{Essential Properties for Uncertainty Attribution}
Adopted from the survey papers \cite{ancona2019gradient,ancona2017towards,nielsen2022robust} for attribution methods of deterministic NNs, some important properties are extended for uncertainty attribution of BDL models. 
\begin{itemize}
    \item \textbf{Implementation Invariance.} The uncertainty attribution methods should assign the same attribution score to the same input for equivalent neural networks, no matter how they are implemented. 
    \item \textbf{Completeness.} The uncertainty score can be fully decomposed into the sum of individual attributions of the input features. 
    \item \textbf{Sensitivity.} The attribution methods should assign zero attribution to the features that will not affect the uncertainty. For two inputs that are different in one feature, this feature should be assigned non-zero attribution if the two inputs lead to different uncertainties. 
    \item  \textbf{Saturation.} Saturation demonstrates a phenomenon in that we assign zero attribution for the regions with zero gradients. The attribution methods should provide tools to avoid saturation. 
    \item \textbf{Linearity.} Denote $f_1, f_2$ as two different BDL models and $M_1(\bm x), M_2(\bm x)$ as the corresponding attribution maps for $\bm x$. The linear combination of the two BDL models is $af_1 +bf_2$, where $a,b\in[0,1]$ and $a+b = 1$. If the linearity is satisfied, the attribution map for $af_1 +bf_2$ is $aM_1(\bm x)+bM_2(\bm x)$.
    \item \textbf{Positivity.} Attribution methods should assign non-negative values to input features. Since features are always imperfect, they should positively contribute to the uncertainty unless they are irrelevant.  
    \item \textbf{Fidelity.} The features with higher attribution scores should be more sensitive to uncertainty change. Through certain changes in the problematic regions, the uncertainty should be significantly reduced. 
\end{itemize}

\subsection{Further Discussion on the Vanilla Extensions of Existing Gradient-based Methods}
\begin{itemize}
    \item \textbf{Grad}. For this method, we use the magnitude of the raw gradients from the uncertainty $U$ to the input $\bm x$, shown in Eq.~\eqref{eq:grad}:  
    \begin{equation}
    \label{eq:grad}
        M_{G}(\bm x) = \left|\frac{\partial U}{\partial \bm x}\right|.
    \end{equation}
    \item \textbf{SmoothGrad}. SmoothGrad tries to smooth the noisy gradients by aggregating from the attributions of various noisy images. Donote $K$ as the number of noisy images we generate through adding Gaussian noises, the attribution map of SmoothGrad is shown in Eq.~\eqref{eq:smoothgrad}:
    \begin{equation}
    \label{eq:smoothgrad}
        M_{SG}(\bm x) = \frac{1}{K} \sum_{k=1}^K M_{G}(\bm x + \mathcal{N}(0,\sigma^2 I))
    \end{equation}
    where $\mathcal{N}(0,\sigma^2 I)$ represents the random noise sampled from the Gaussian distribution with 0 mean and covariance matrix $\sigma^2 I$. $\sigma$ is a hyperparameter and $I$ is the identity matrix. 
    \item \textbf{FullGrad}. The FullGrad method calculates the attribution map $M_{FG}(\bm x)$ by considering both the gradient of the uncertainty measure $U$ with respect to the input $\bm x$ (i.e., $\frac{\partial U}{\partial \bm x}$) and the gradient of $U$ with respect to the bias variable $\bm b_{l}$ in every convolutional or fully-connected layer $l$ (i.e., $\frac{\partial U}{\partial \bm b_l}$). This aggregation is mathematically expressed in Eq.~\eqref{eq:fullgrad}:
\begin{equation}
\label{eq:fullgrad}
    M_{FG}(\bm x) = \psi\left(\left|\frac{\partial U}{\partial \bm x} \odot \bm x \right| + \sum_{l}  \left|\frac{\partial U}{\partial \bm b_l}\odot \bm b_{l}\right|\right)
\end{equation}
where $\odot$ is the element-wise product and $|\cdot|$ returns the absolute values. $\psi$ is a post-processing function for normalizing and rescaling the gradients.
\item \textbf{Itegrated Gradient (IG)}. Integrated gradient method creates a path integral from a reference image $\bm x_0$ to $\bm x$, shown in Eq.~\eqref{eq:IG}:
\begin{equation}
    \label{eq:IG}
    M_{IG}(\bm x) = (\bm x - \bm x_0) \odot \int_{0}^1 \frac{\partial U(\bm x_0 +\alpha (\bm x - \bm x_0))}{\partial \bm x} d\alpha .
\end{equation}
Since IG requires a reference image $\bm x_0$ and the attribution results highly depend on the difference between the reference image and the original image, various extensions are proposed, leading to Blur IG \cite{xu2020attribution} and Guided IG \cite{kapishnikov2021guided}. 
\end{itemize}
Based on the survey papers \cite{ancona2019gradient,ancona2017towards,nielsen2022robust}, we briefly summarize the properties satisfied by the aforementioned approaches in Table \ref{tab:property}. In the next section, we will show the theoretical analysis of our proposed method.

\begin{table*}[ht]\small
	\caption{The properties of the selected gradient-based attribution methods. The ``Yes" in saturation means the attribution method has tools to avoid zero attribution for zero-gradient regions. ``*" means the property depends on specific architectures or the chosen layers.}
	\label{tab:property}
	\centering
\begin{tabular}{|l|ccccccc|l|}
\hline
	\multirow{2}{*}{Method}  & \multicolumn{7}{c|}{Properties} \\
 	\cline{2-8}
 &Implementation Invariance &Completeness &Sensitivity &Saturation &Linearity &Positivity &Fidelity \\
\hline  
\multirow{1}{*}{Grad} &Yes &No &Yes &No &No &Yes &No \\
\multirow{1}{*}{SmoothGrad} &Yes &No &Yes &No &No &Yes &Yes \\
\multirow{1}{*}{FullGrad} &Yes* &Yes &Yes &Yes &No &Yes &Yes \\
\multirow{1}{*}{IG} &Yes &Yes &Yes &Yes &Yes &No &Yes \\
\hline
\end{tabular}
\end{table*}

\subsection{Special Properties of UA-Backprop}
\proposition{UA-Backprop always satisfies the completeness property.}
\proof{Based on Algorithm 1 of the main body of the paper, the uncertainty attribution map generated by our proposed method is shown in Eq.~\eqref{eq:attribution_ours}:
\begin{equation}
    \label{eq:attribution_ours}
    M(\bm x) = \frac{1}{S} \sum_{s=1}^S \sum_{i=1}^C U_{z_i}^s M_i^s(\bm x)
\end{equation}
where $M_i^s(\bm x)$ is the normalized relevance map showing the essential regions of $\bm x$ that contribute to $\bm z_i^s$. $U_{ z_i}^s$ is the uncertainty attribution of $\bm z_i^s$ received from $\bm g^s$. By taking the sum of $M(\bm x)$ over all the elements,
\begin{equation}
    \begin{split}
        &\sum_{(u,v)} M(\bm x)[u,v] \\
        &= \frac{1}{S} \sum_{s=1}^S \sum_{i=1}^C U_{z_i}^s \sum_{(u,v) }M_i^s(\bm x)[u,v] \\
        &= \frac{1}{S} \sum_{s=1}^S \sum_{i=1}^C U_{z_i}^s  = \frac{1}{S} \sum_{s=1}^S \sum_{i=1}^C \sum_{j=1}^C c_{g_j^s \rightarrow z_i^s} U_{g_j} \\
        &=\frac{1}{S} \sum_{s=1}^S \sum_{j=1}^C (\sum_{i=1}^C c_{g_j^s \rightarrow z_i^s}) U_{g_j} \\
        &=\frac{1}{S} \sum_{s=1}^S \sum_{j=1}^C U_{g_j}  = \frac{1}{S} \sum_{s=1}^S U = U\\
    \end{split}
\end{equation}}

By incorporating the FullGrad method into the attribution proposed backpropagation framework for the path $\bm z \rightarrow \bm x$, our method is able to satisfy several crucial properties. It should be noted that the fulfillment of these properties is primarily contingent on the choice of backpropagation method employed for $\bm z \rightarrow \bm x$, as the attribution propagation from $U \rightarrow \bm g$ and $\bm g \rightarrow \bm z$ does not involve neural network parameters. In the case of UA-Backprop + FullGrad, our method is able to achieve completeness, sensitivity, saturation, positivity, and fidelity.

\section{Implementation Details and Experiment Settings}
In this section, we will discuss the implementation details of the proposed method and provide further information about the experiment settings. 
\subsection{Implementation Details and Training Hyperparameters}
\subsubsection{Model Architecture}
As described in Sec.~5 of the main body of the paper, we adopt the deep ensemble method to estimate the uncertainty. Specifically, we train an ensemble of five models for each dataset with different initialization seeds. Common data augmentation techniques, such as random cropping and horizontal flipping, are applied to C10, C100, and SVHN datasets. Our experiments are conducted on an RTX2080Ti GPU using PyTorch. The model architecture and hyperparameters used in our experiments are detailed below.
\begin{itemize}
    \item \textbf{MNIST}. We use the architecture: Conv2D-Relu-Conv2D-Relu-MaxPool2D-Dropout-Dense-Relu-Dropout-Dense-Softmax. Each convolutional layer contains 32 convolution filters with $4\times4$ kernel size. We use a max-pooling layer with a $2\times2$ kernel, two dense layers with 128 units, and a dropout probability of 0.5. The batch size is set to 128 and the maximum epoch is 30. We use the SGD optimizer with a learning rate of $0.1$ and momentum of $0.9$. 
    \item \textbf{C10}. For the C10 dataset, we employ ResNet18 as the feature extractor, followed by a single fully-connected layer for classification. We use the stochastic gradient descent (SGD) optimizer with an initial learning rate of $0.1$ and momentum of $0.9$. The maximum number of epochs is set to 100, and we reduce the learning rate to $0.01$, $0.001$, and $0.0001$ at the 30th, 60th, and 90th epochs, respectively. The batch size is set to 128.
    \item \textbf{C100}. For the C100 dataset, we use the same model architecture as in C10, with ResNet18 as the feature extractor and a single fully-connected layer for classification. We adopt the SGD optimizer with an initial learning rate of $0.1$ and momentum of $0.9$. The maximum number of epochs is set to 200, and we decrease the learning rate to $0.01$, $0.001$, and $0.0001$ at the 60th, 120th, and 160th epochs, respectively. The batch size is set to 64.
    \item \textbf{SVHN}. We use the same architecture as MNIST. The batch size is set to 64 and the maximum epoch is 50. We use the SGD optimizer with a learning rate of $0.1$ and momentum of $0.9$. The learning rate is decreased to 0.01 and 0.001 at the 15th and 30th epochs. 
\end{itemize}
\subsubsection{Implementation of the Attribution Approaches}
\begin{itemize}
    \item \textbf{Ours}. Regarding the MNIST dataset, we set $\tau_1$ and $\tau_2$ to $0.08$ and $0.3$ respectively, whereas for C10, C100, and SVHN, we set $\tau_1$ and $\tau_2$ to $0.55$ and $0.02$. The hyperparameters are different since MNIST contains only grayscale images, while the other datasets consist of colorful images. We utilize the FullGrad method, which is an internal part of the UA-Backprop for $\bm z \rightarrow \bm x$, and we refer to the implementation available at \url{https://github.com/idiap/fullgrad-saliency} with the default hyperparameters.
    \item \textbf{Grad}. We use the Torch.autograd to directly compute the gradient from the uncertainty score and the input. 
    \item \textbf{SmoothGrad}. Based on Eq.~\eqref{eq:smoothgrad}, we use $K=50, \sigma = 0.1$ to smooth the gradients.
    \item \textbf{FullGrad}. We use the implementation in \url{https://github.com/idiap/fullgrad-saliency} as a basis and extend it to the uncertainty attribution analysis by computing the full gradients from the uncertainty score to the input. We utilize the default hyperparameters. 
    \item \textbf{Blur IG and IG}. We follow \url{https://github.com/Featurespace/uncertainty-attribution} for the uncertainty-adapted versions of the Blur IG and IG. The number of path integrations used for Blur IG and IG is set to 100. We use the white starting image for IG. 
    \item \textbf{CLUE and $\delta$-CLUE}. For CLUE and $\delta$-CLUE, a two-stage process is performed where we first train two variational autoencoders (VAEs). Specifically, for the MNIST dataset, the VAE implementation follows that of \url{https://github.com/lyeoni/pytorch-mnist-VAE/blob/master/pytorch-mnist-VAE.ipynb}. Meanwhile, for C10, C100, and SVHN datasets, we utilize the implementation of \url{https://github.com/SashaMalysheva/Pytorch-VAE}, with the same model architectures and the default hyperparameters. The output layer of the aforementioned implementation is modified to use a sigmoid activation function for the binary cross-entropy loss. Once the VAEs are trained, we apply the CLUE and $\delta$-CLUE methods to learn a modified image for each test data, where the uncertainty loss and the reconstruction loss are weighted equally. We use Adam optimizer with a learning rate of 0.01 and set the maximum iteration to 500 with an early stop criteria based on an L1 patience of $1e-3$.
\end{itemize}
\subsection{Experiment Settings}
\begin{table*}[h]
\fontsize{9}{12}\selectfont
	\caption{IoU $\uparrow$ and ADA $\uparrow$  for anomaly detection for various datasets. The bold values indicate the best performance}
	\label{tab:anomaly_detection}
	\centering
\begin{tabular}{|l|cccccccc|l|}
\hline
	\multirow{2}{*}{Method} 
	& \multicolumn{2}{c}{C10} & \multicolumn{2}{c}{C100} & \multicolumn{2}{c}{SVHN} & \multicolumn{2}{c|}{Avg. Performance} \\
 &IoU &ADA &IoU &ADA &IoU &ADA &IoU &ADA\\
\hline  
\multirow{1}{*}{Ours}  &\textbf{0.353}&\textbf{0.285}&\textbf{0.363}&\textbf{0.375}&\textbf{0.217}&\textbf{0.124}&\textbf{0.311}&\textbf{0.261}\\
\multirow{1}{*}{Grad}  &0.141&0.090&0.167&0.135&0.198&0.096&0.169&0.107\\
\multirow{1}{*}{SmoothGrad}  &0.321&0.260&0.316&0.245&0.212&0.114&0.283&0.206\\
\multirow{1}{*}{FullGrad}  &0.341&\textbf{0.285}&0.320&0.295&0.206&0.114&0.289&0.231\\
\multirow{1}{*}{IG}  &0.171&0.090&0.170&0.105&0.139&0.052&0.160&0.082\\
\multirow{1}{*}{Blur IG}  &0.182&0.125&0.318&0.290&0.150&0.078&0.217&0.164\\
\multirow{1}{*}{CLUE}  &0.253&0.210&0.208&0.180&0.115&0.042&0.192&0.114\\
\multirow{1}{*}{$\delta-$CLUE}  &0.248&0.240&0.229&0.220&0.105&0.044&0.194&0.168\\
\hline
\end{tabular}\vspace{-3mm}
\end{table*}
\subsubsection{Blurring Test}
In Sec.~5 of the main context, we examine the performance of the epistemic uncertainty maps in a blurring test. In this test, the key hyperparameter is the standard deviation $\sigma$ of the Gaussian filter. However, using a fixed $\sigma$ would be unfair since a small $\sigma$ would have no impact on the image, while a large $\sigma$ would cause the blurred images to be out-of-distribution. Different images may require varying degrees of blurriness to reduce uncertainty appropriately. Therefore, we perform an individual search for $\sigma$ for each image, ensuring that the blurred image has the minimum uncertainty. The search range is from 0 to 20, with a step of 0.2. As our proposed method aims to identify problematic regions by analyzing uncertain images, we focus on the top 500 images with the highest epistemic uncertainty for the blurring test evaluation. Note that for MNIST dataset, only the top 100 uncertain images are selected for evaluation since most of the images have a good quality with low uncertainty. For each metric, the median value is reported considering that some blurred images could be out-of-distribution with increased uncertainty.

\subsubsection{Uncertainty Mitigation With Attention Mechanism}
In this study, we aim to improve model performance by using pre-generated uncertainty maps as attention to mitigate uncertainty. Following Eq. (12) of the main body of the paper, the uncertainty attribution map $M(\bm x)$ is first normalized using an element-wise softmax function and then used for constructing the attention $A(\bm x)$. We use bilinear interpolation to rescale $A(\bm x)$ to the size of the hidden feature maps. We then do an element-wise product of $(1+\alpha A(\bm x))$ with the hidden features, where $\alpha$ is a positive real number that controls the strength of the attention. We choose $\alpha =0.2$ across all datasets and adding $1$ is to keep the information of the regions with low importance to ensure no knowledge loss. In the main experiment, we use the epistemic uncertainty maps, while an ablation study for using aleatoric and total uncertainty maps as attention is provided in Appendix D.2.3. To evaluate model robustness, we retrain the model with the attention mechanism under limited data and test on the original testing dataset. With limited data, there is no need for applying complex models. Hence, we use the CNN-based models for all the datasets. The model architecture is Conv2D-Relu-Conv2D-Relu-MaxPool2D-Dropout-Dense-Relu-Dropout-Dense-Relu-Dense-Softmax. Each convolutional layer contains 32 convolution filters with $4\times4$ kernel size. We use a max-pooling layer with a $2\times2$ kernel, several dense layers with 128 units, and a dropout probability of 0.5. The maximum training epoch is 120 and the batch size is 128. We use the SGD optimizer with an initial learning rate of $0.1$ and momentum of $0.9$. The learning rate is decreased at the 30th, 60th, and 90th epoch with a decay rate of $0.2$. Additional results for different experiment settings can be found in Appendix D.2.

\section{Anomaly Detection}

\indent In this section, we employ our method to conduct anomaly detection by leveraging the known ground-truth problematic regions. Specifically, we substitute one patch of each testing image with a random sample from the training data at an identical location. Despite the modified patch still being marginally in-distribution, it mismatches with the remaining regions, creating the ground-truth problematic regions. We perform a quantitative assessment of the efficacy of our proposed method in detecting these anomaly patches. 

\indent The experimental evaluation is conducted on three datasets, namely C10, C100, and SVHN. MNIST is excluded from the comparison due to its grayscale nature. To generate the ground-truth problematic regions, we randomly modify a 10 by 10 patch in each testing image by replacing it with a sample from the training data distribution at the same location. Out of the resulting modified images, we select 200 images that exhibit the largest increase in uncertainty compared to the original images, indicating the most problematic areas. Then the epistemic uncertainty maps are generated, based on which, we predict the troublesome regions by fitting a 10 by 10 bounding box that has the highest average attribution score. It is worth noting that we use a brute-force method to identify the predicted 10 by 10 patch. The predicted bounding boxes are compared with the ground-truth counterparts using Intersection over Union (IoU) and anomaly detection accuracy (ADA). The IoU is calculated by dividing the area of the overlap by the area of union, while the detection accuracy is the percentage of images with IoU greater than 0.5.
\begin{figure}[h]
    \centering
    \includegraphics[width=1.0\linewidth]{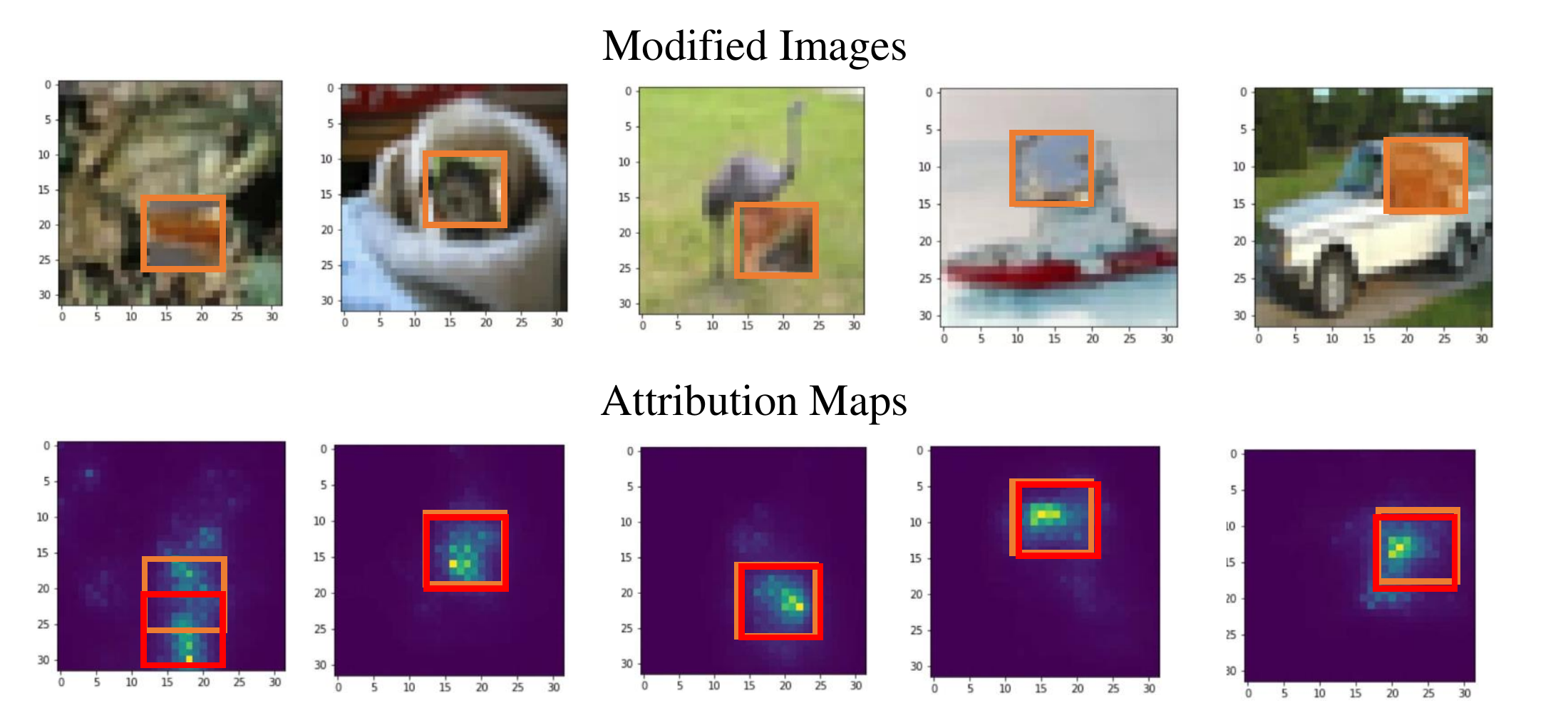}
    \caption{The anomaly detection examples. The red bounding boxes represent the predicted problematic regions while the orange bounding boxes are the ground truth.}
    \label{fig:anomaly_example}
\end{figure}\\
As shown in Figure \ref{fig:anomaly_example}, the predicted problematic bounding boxes are well-matched with the ground truth, indicating the method's capability to accurately identify anomalous regions. The quantitative evaluation in Table \ref{tab:anomaly_detection} reveals that UA-Backprop outperforms other baselines, especially for Grad, IG, Blur IG, CLUE, and $\delta$-CLUE. These baselines perform poorly in detecting anomalous regions, which may be attributed to their limited ability to identify continuous problematic regions (i.e., the 10 by 10 patches), as they tend to detect only scattered locations. %
\begin{table*}[ht]\small
	\caption{Acc (\%) and NLL for uncertainty mitigation evaluation of varying number of training samples $N$ on MNIST and C10 datasets. The results are aggregated over 5 independent runs. }
	\label{tab:vary_n}
	\centering
\begin{tabular}{|l|cccccccccccc|l|}
\hline
	\multirow{3}{*}{Method} & \multicolumn{12}{|c|}{MNIST}\\
	\cline{2-13}
	& \multicolumn{2}{c}{$N=200$} & \multicolumn{2}{c}{$N=500$} & \multicolumn{2}{c}{$N=1000$}& \multicolumn{2}{c}{$N=1500$} & \multicolumn{2}{c}{$N=2000$} & \multicolumn{2}{c|}{Avg. Performance}\\
 &ACC &NLL &ACC &NLL &ACC &NLL &ACC &NLL &ACC &NLL &ACC &NLL\\
\hline  
\multirow{1}{*}{Ours}  &\textbf{85.86} &\textbf{0.461} &91.95 &\textbf{0.287} &\textbf{95.65} &0.186 &\textbf{96.43} &0.161&\textbf{96.72}&0.152 &\textbf{93.32}&\textbf{0.249} \\
\multirow{1}{*}{Grad}  &85.02 &0.490 &91.35 &0.302 &94.89 &0.192 &95.76 &0.176&96.47&0.159 &92.70&0.264\\
\multirow{1}{*}{SmoothGrad}  &85.38 &0.480 &90.68 &0.324 &95.15 &0.188 &95.97 &0.171&96.35&0.159 &92.71&0.264\\
\multirow{1}{*}{FullGrad}  &84.75 &0.503 &91.39 &0.300 &95.23 &\textbf{0.175} &95.98 &\textbf{0.153}&96.44&\textbf{0.142} & 92.76 &0.255\\
\multirow{1}{*}{IG}  &82.66&0.563&\textbf{91.98}&0.350&94.94&0.220&95.71&0.190&96.34&0.162&92.33&0.297\\
\multirow{1}{*}{Blur IG}  &85.34 &0.485 &91.57 &0.288 &95.04 &0.184 &96.02 &0.155&96.48&0.145 &92.89&0.252\\
\multirow{1}{*}{Non-attention}  &84.64 &0.524 &90.78 &0.358 &95.01 &0.221 &95.94 &0.189&96.29&0.172&92.55&0.293 \\
\hline
\end{tabular}\\
\vspace{+1mm}
\begin{tabular}{|l|cccccccccccc|l|}
\hline
	\multirow{3}{*}{Method} & \multicolumn{12}{|c|}{C10}\\
	\cline{2-13}
	& \multicolumn{2}{c}{$N=1000$} & \multicolumn{2}{c}{$N=2000$} & \multicolumn{2}{c}{$N=3000$}& \multicolumn{2}{c}{$N=4000$} & \multicolumn{2}{c}{$N=5000$} & \multicolumn{2}{c|}{Avg. Performance}\\
 &ACC &NLL &ACC &NLL &ACC &NLL &ACC &NLL &ACC &NLL &ACC &NLL\\
\hline  
\multirow{1}{*}{Ours}  & \textbf{36.48} &\textbf{1.768} &\textbf{49.25} &\textbf{1.454} &52.17 &1.377&\textbf{56.92}&\textbf{1.255}&57.64&1.222 &\textbf{50.49}& \textbf{1.415}\\
\multirow{1}{*}{Grad}  & 31.47 &1.945 &47.35 &1.472 &51.62 &1.374 &46.91&1.508&52.85 &1.322 &46.04&1.524\\
\multirow{1}{*}{SmoothGrad}  &31.73  &1.944 &42.72 &1.943 &48.15 &2.482 &47.96&2.342&48.02&2.435&43.72&2.229\\
\multirow{1}{*}{FullGrad}  &32.53 &1.920 &46.67&1.485&51.46&1.371&54.57&1.290&53.70&1.297&47.79&1.473\\
\multirow{1}{*}{IG}  &34.43&1.829&47.96&1.472&\textbf{53.32}&\textbf{1.349}&56.37&1.263&\textbf{58.41}&\textbf{1.200}&50.10&1.423\\
\multirow{1}{*}{Blur IG}  & 31.96 &1.932 &46.38 &1,495 &52.11 &1.364 &52.48&1.335&54.71&1.277 &47.53&1.481\\
\multirow{1}{*}{Non-attention} & 31.58 &1.922 &46.57 &1.490 &51.25 &1.378 &54.89&1.281&55.11&1.265 &47.88&1.467 \\
\hline
\end{tabular}

\end{table*}
\section{Ablation Studies and Further Analysis}
\subsection{Efficiency Evaluation}
In this section, we present a theoretical efficiency analysis of gradient-based methods for generating uncertainty maps. We define the runtime of a single backpropagation as $O(1)$. Our proposed method, along with Grad and FullGrad, can generate the maps within a single backpass, resulting in a runtime of $O(1)$. However, SmoothGrad, IG, and Blur IG require multiple backward passes for attribution analysis, with runtimes of $O(T)$ where $T$ represents the number of backward iterations. For SmoothGrad, the value of $T$ depends on the number of noisy images used for aggregation, while for the IG-based method, $T$ is based on the number of samples generated to approximate the path integral. The CLUE-based methods necessitate solving an optimization problem per input to obtain a modified image for reference, which further extends their runtimes. In Table \ref{tab:efficiency}, we provide the empirical results on the runtime required for each baseline to attribute a single image, demonstrating that our proposed method outperforms various baselines in terms of computational efficiency.
\begin{table}[h]\small
    \centering
    \caption{Runtime (s) for attributing one image.}
    \label{tab:efficiency}
     \vspace{-3mm}
\begin{tabular}{|l|c c c c|l|}
   \hline
   Dataset/Method  &Ours &Blur-IG &SmoothGrad & CLUE\\
   \hline
    MNIST &\textbf{0.34}&3.39&3.06&6.93 \\
    C10 &\textbf{0.46} &4.06&3.59&18.43 \\
   \hline
\end{tabular} \vspace{-4mm}
\end{table}\\
\subsection{Different Experiment Settings for Uncertainty Attention Mitigation}
\subsubsection{Varying Number of Training Samples}
In this section, we present the results of a study in which we investigate the effect of varying the number of training samples on the MNIST and C10 datasets in the context of the retaining with the attention mechanism. The experimental outcomes are reported in Table \ref{tab:vary_n}. We observe that our proposed method consistently outperforms other methods when the training data is limited, as evidenced by the improved testing accuracy and NLL. We also find that adding attention to the training of the C10 dataset may not be beneficial for some methods, possibly due to the noisy gradients.

\begin{table*}[h]\small
	\caption{Acc (\%) and NLL for uncertainty mitigation evaluation of varying $\alpha$ on MNIST and C10 datasets for our proposed method. We randomly select 500 and 1000 training samples for MNIST and C10, respectively. The results are aggregated over 5 independent runs. $\alpha = 0.2$ is used for the main body of the paper.}
	\label{tab:vary_alpha}
	\centering
\begin{tabular}{|l|cccccc|l|}
\hline
	\multirow{3}{*}{$\alpha=$} & \multicolumn{6}{|c|}{Dataset}\\
	\cline{2-7}
	& \multicolumn{2}{c}{MNIST} & \multicolumn{2}{c}{C10} &  \multicolumn{2}{c|}{Avg. Performance}\\
 &ACC &NLL &ACC &NLL &ACC &NLL\\
\hline  
\multirow{1}{*}{0.0}  &90.78&0.358&31.62&1.921 &61.20&1.140 \\
\multirow{1}{*}{0.2}  &91.95 &0.287 &36.48&1.768 &64.22&1.028 \\
\multirow{1}{*}{0.4}  &91.62 &0.329 &35.22 &1.806 &63.42&1.068  \\
\multirow{1}{*}{0.6}  &91.98 &0.320 &35.73 &1.793  &63.86&1.057\\
\multirow{1}{*}{0.8}  &92.07 &0.297 &\textbf{38.39} &\textbf{1.735} &65.23&1.016\\
\multirow{1}{*}{1.0}  &92.17 &0.299 &36.42 &1.779  &64.30&1.068\\
\multirow{1}{*}{1.2}  &92.28 &0.285 &37.59 &1.750 &\textbf{64.94} &1.018\\
\multirow{1}{*}{1.4}  &91.86 &0.307 &38.00 &1.737 &64.93&1.022\\
\multirow{1}{*}{1.6}  &91.99 &0.295 &36.24 &1.782 &64.12&1.038  \\
\multirow{1}{*}{1.8}  &\textbf{92.52} &\textbf{0.269} &36.52 &1.760 &64.52&1.015 \\
\multirow{1}{*}{2.0}  &92.51 &\textbf{0.269} &37.77 &1.743 &65.14 &\textbf{1.006} \\
\hline
\end{tabular}\vspace{-3mm}
\end{table*}

\begin{table*}[h]\small
	\caption{Acc (\%) and NLL for uncertainty mitigation evaluation with different kinds of uncertainty maps.}
	\label{tab:vary_type}
	\centering
\begin{tabular}{|l|cccc|l|}
\hline
	\multirow{3}{*}{Uncertainty} & \multicolumn{4}{|c|}{Dataset}\\
	\cline{2-5}
	& \multicolumn{2}{c}{MNIST} & \multicolumn{2}{c|}{C10} \\
 &ACC &NLL &ACC &NLL\\
\hline  
\multirow{1}{*}{Epistemic} &\textbf{91.95} &\textbf{0.287} &36.48&1.768  \\
\multirow{1}{*}{Aleatoric} &91.94 &0.315 &\textbf{37.38} &\textbf{1.761} \\
\multirow{1}{*}{Total}  &91.60 &0.330 &35.14 & 1.810\\
\hline
\end{tabular}\vspace{-5mm}
\end{table*}

\begin{table*}[h]\small
	\caption{Mitigation results (ACC $\uparrow$,NLL $\downarrow$) for MNIST and C10. The comparison is conducted for input-space attention and latent-space attention for uncertainty mitigation.}
	\label{tab:input-space}
	\centering
 \vspace{-3mm}
\begin{tabular}{|l|cccccc|l|}
\hline
	\multirow{2}{*}{Method} 
	& \multicolumn{2}{c}{MNIST} & \multicolumn{2}{c}{C10} & \multicolumn{2}{c|}{Average}\\
 &ACC &NLL  &ACC &NLL &ACC &NLL\\
\hline
\multirow{1}{*}{Ours-latent}  &\textbf{0.920}&0.287&0.365&1.768 &0.642&1.028\\
\multirow{1}{*}{Ours-input}  &0.919&\textbf{0.284}&\textbf{0.376}&\textbf{1.742} &\textbf{0.648}&\textbf{1.013}\\
\hline
\end{tabular} 
\end{table*} 

\begin{table*}[h]
\fontsize{8.5}{11}\selectfont
	\caption{MURR and AUC-URR (AUC) of the blurring test for our proposed method with different hyperparameters. The number of blurring pixels is $2\%$ or $5\%$ of the total pixels. The first row shows the hyperparameters used for displaying the main results. The studies are conducted on SVHN dataset.}
	\label{tab:vary_hyper}
	\centering
\begin{tabular}{|l|cccc|cccc|cccc|l|}
\hline
	\multirow{3}{*}{Hyperparameter} & \multicolumn{4}{|c|}{Dataset - SVNH } & \multicolumn{4}{|c|}{Dataset - C10 }& \multicolumn{4}{|c|}{Dataset - C100 }\\
	\cline{2-13}
	& \multicolumn{2}{c}{2\%} & \multicolumn{2}{c|}{5\%} & \multicolumn{2}{c}{2\%} & \multicolumn{2}{c|}{5\%}& \multicolumn{2}{c}{2\%} & \multicolumn{2}{c|}{5\%}\\
 &MURR &AUC &MURR &AUC &MURR &AUC &MURR &AUC &MURR &AUC &MURR &AUC\\
\hline  
\multirow{1}{*}{$\tau_1 = 0.55, \tau_2 = 0.02$} &0.625  &0.526 &0.758&0.407 &\textbf{0.629}&0.664&0.848&0.484 &\textbf{0.195}&0.901&0.302&0.821\\
\multirow{1}{*}{$\tau_1 = 0.50, \tau_2 = 0.02$} &0.603 &0.550 &0.739&0.419 &0.622&0.664&0.848&0.496 &0.194&\textbf{0.900}&\textbf{0.304}&0.821\\
\multirow{1}{*}{$\tau_1 = 0.60, \tau_2 = 0.02$}  &0.607 &0.540 &0.732 &0.407 &0.626&0.666&0.850&0.489 &0.194&0.901&0.303&0.821\\
\multirow{1}{*}{$\tau_1 = 0.65, \tau_2 = 0.01$}  &\textbf{0.645} &0.518 &\textbf{0.771}  &0.397 &0.617&0.666&0.848&0.491&0.194&0.901&0.298&\textbf{0.820}\\
\multirow{1}{*}{$\tau_1 = 0.70, \tau_2 = 0.02$}  &0.595 &0.545 &0.747  &0.419 &0.624&\textbf{0.660}&\textbf{0.854}&\textbf{0.480}&0.194&0.901&0.302&0.821\\
\multirow{1}{*}{$\tau_1 = 0.55, \tau_2 = 0.03$}  &0.635 &\textbf{0.509} &0.758  &\textbf{0.387} &0.603&0.690&0.848&0.510 &0.194&0.905&0.296&0.831\\
\multirow{1}{*}{$\tau_1 = 0.55, \tau_2 = 0.04$}  &0.608 &0.562 &0.761  &0.406 &0.592&0.682&0.829&0.508 &0.190&0.903&0.294&0.835 \\
\hline
\end{tabular}
\end{table*}

\subsubsection{Varying Hyperparameters}
In this section, we investigate the impact of the attention weight coefficient, denoted by $\alpha$, on the performance of our proposed method for MNIST and C10 datasets. We vary $\alpha$ from 0 to 2 with a step of 0.2 and present the results in Table \ref{tab:vary_alpha}. Our proposed method consistently outperforms the plain training without attention ($\alpha=0$) as we vary $\alpha$. In this study, we set $\alpha$ to a minimum value of 0.2. Remarkably, even a small value of $\alpha$ leads to a significant improvement. Furthermore, larger values of $\alpha$ progressively accentuate the informative regions, resulting in better performance, as evidenced by the improved results for $\alpha=1.8, 2$ on MNIST and $\alpha=1.2,1.4$ on C10. Considering the stochastic nature of the training process, we note that the model's performance is insensitive to $\alpha$ within a certain range.

\begin{table*}[ht]\small
	\caption{MURR and AUC-URR (AUC) of the blurring test for our proposed method with different approachs for $\bm z \rightarrow \bm x$. The number of blurring pixels is $2\%$ or $5\%$ of the total pixels. The studies are conducted on MNIST and SVHN datasets.}
	\label{tab:different_z}
	\centering
\begin{tabular}{|l|cccc|cccc|l|}
\hline
	\multirow{3}{*}{Method} & \multicolumn{4}{|c|}{MNIST} & \multicolumn{4}{|c|}{SVHN}\\
	\cline{2-9}
	& \multicolumn{2}{c}{2\%} & \multicolumn{2}{c|}{5\%} & \multicolumn{2}{c}{2\%} & \multicolumn{2}{c|}{5\%}\\
 &MURR &AUC &MURR &AUC &MURR &AUC &MURR &AUC\\
\hline  
\multirow{1}{*}{UA-Backprop + FullGrad} &0.648  &0.667 &\textbf{0.850}&0.445 &\textbf{0.625} &\textbf{0.526} &\textbf{0.758}&\textbf{0.407}  \\
\multirow{1}{*}{UA-Backprop + Grad} &0.519 &0.714 &0.720 &0.532 &0.611 &0.543 &0.712 &0.451 \\ 
\multirow{1}{*}{UA-Backprop + InputGrad} &\textbf{0.673}  &\textbf{0.618} &0.826&\textbf{0.413} &0.549 &0.598&0.702&0.445  \\
\multirow{1}{*}{UA-Backprop + IG} &0.611  &0.641 &0.795&0.439 &0.529 &0.618 &0.703&0.456 \\
\hline
\end{tabular}
\end{table*}

\subsubsection{Aleatoric and Total Uncertainty Map}
In this study, we explore the use of alternative uncertainty maps, namely aleatoric and total uncertainty maps, in place of epistemic uncertainty maps as the attention mechanism. Table \ref{tab:vary_type} presents a comparison of model performance using different types of uncertainty maps. While all maps exhibit a similar accuracy on the MNIST dataset, utilizing the epistemic uncertainty maps results in better fitting based on NLL. On the C10 dataset, the aleatoric uncertainty maps yield slightly better performance in both ACC and NLL. Since aleatoric uncertainty captures input noise, the aleatoric uncertainty maps can strengthen the regions with less noise and may benefit when the input imperfections result mainly from input noise. The superior results for aleatoric uncertainty maps on the C10 dataset may be due to the fact that the C10 dataset is noisier than the MNIST dataset. 

\subsubsection{Input/Latent-space Attention for Uncertainty Mitigation}
Table \ref{tab:input-space} presents our experimental results using UA maps as input-space attention. The weighted inputs $A(\bm x) \odot \bm x$ are obtained by using $A(\bm x)$ as input attention. We then use the weighted inputs to retrain the model under the same experimental conditions as described in Appendix B.2.2. Our results demonstrate that using UA maps as input-space attention yields similar performance compared to the results obtained through latent-space experiments.

\vspace{+3mm}
\subsection{Hyperparameter Sensitivity of Our Proposed Method}
The temperatures $\tau_1$ and $\tau_2$ used in the normalization functions are crucial hyperparameters in our proposed method. It is necessary to perform normalization in the intermediate steps to ensure the satisfaction of the completeness property. By choosing appropriate values for $\tau_1$ and $\tau_2$, we aim to avoid uniform or overly sharp coefficients. It is essential to avoid setting $\tau_1$ and $\tau_2$ too small or too large, as this would result in uniform or extreme scores. In this section, we show some blurring test results for SVHN, C10, and C100 datasets to evaluate the sensitivity of $\tau_1, \tau_2$ within a certain range. In Table \ref{tab:vary_hyper}, the first row shows the hyperparameters used for the experiments of the main body of the paper. We can observe that the performance varies slightly by choosing different hyperparameters within certain ranges. During experiments, we tune $\tau_1, \tau_2$ on C10 dataset and use the same hyperparameters ($\tau_1=0.55, \tau_2=0.02$) for all other datasets with color images. Since MNIST contains only gray-scale images, we use a different set of hyperparameters, i.e., $\tau_1=0.08, \tau_2=0.3$. It is worth noting that the cross-dataset results are insensitive to the variations of $\tau_1, \tau_2$ within certain ranges. Tuning different $\tau_1, \tau_2$ for different datasets can further improve the performance.

\subsection{Different Methods for the Path $z \rightarrow x$}
As described in Sec.~3 of the main paper, the UA-Backprop method has the potential to serve as a general framework for utilizing advanced gradient-based techniques to investigate the path from $\bm z$ to $\bm x$. By exploring the path $z_i^s \rightarrow \bm x$, we obtain the relevance map $M_i^s(\bm x)$, which highlights the crucial regions of $\bm x$ for $z_i^s$, as presented in Eq.~(10) of the main paper. Although we use the FullGrad method as our primary approach, other gradient-based techniques can also be employed within the UA-Backprop framework. As a simple baseline, UA-Backprop + Grad uses
\begin{equation}
     M_i^s(\bm x) = \psi\left(\frac{\partial z_i^s}{\partial \bm x}\right)
\end{equation}
where $\psi$ is a softmax function with temperature $\tau_2$, similar to UA-Backprop + FullGrad. However, the raw gradients could be noisy, and advanced gradient-based methods could be used. For example, UA-Backprop + InputGrad uses
\begin{equation}
 M_i^s(\bm x) = \psi\left(\bm x  \odot \frac{\partial z_i^s}{\partial \bm x}\right)
\end{equation}
where the input image is used to smooth the gradients. We can also use the integrated gradient (IG) method, which is an extension of InputGrad by creating a path integral from a reference image $\bm x_0$ to input $\bm x$. For UA-Backprop + IG,
\begin{equation}
\begin{split}
 & M_i^s(\bm x)  \\
 &= \psi\left((\bm x - \bm x_0) \odot \int_{0}^1 \frac{\partial z_i(\bm x_0 +\alpha (\bm x - \bm x_0), \boldsymbol \theta^s)}{\partial \bm x} d\alpha \right)
 \end{split}
\end{equation}
where $\bm x_0$ could be a black or white image as the reference. In this section, we provide an ablation study of using different gradient-based methods for the path $\bm z \rightarrow \bm x$. The blurring test evaluations are provided in Table \ref{tab:different_z} for MNIST and SVHN datasets. The first row "UA-Backprop + FullGrad" represents the method shown in the main body of the paper. By using other methods for the path $\bm z \rightarrow \bm x$, we can also achieve considerable results. For example, UA-Backprop + InputGrad can achieve some improvements for the MNIST dataset. In short, our proposed method can be a general framework combining the recent development of other gradient-based methods for deterministic NNs. 
\begin{table}[h]
\fontsize{7.8}{11.0}\selectfont
	\caption{Attribution results (MURR $\uparrow$, AUC-URR $\downarrow$).}
	\label{tab:single}
	\centering
 \vspace{-3mm}
\begin{tabular}{|l|cccc|l|}
\hline
	\multirow{2}{*}{Method} 
	& \multicolumn{2}{c}{MNIST (\%2)} & \multicolumn{2}{c|}{C10 (\%2)} \\
 &MURR &AUC-URR  &MURR &AUC-URR  \\
\hline
\multirow{1}{*}{Ours-Ensemble-5}  &\textbf{0.648} &\textbf{0.667} &\textbf{0.629} &\textbf{0.664}\\
\multirow{1}{*}{Ours-Ensemble-1}  &0.425 &0.828 &0.506 &0.710 \\
\multirow{1}{*}{Ours-LA}  &0.487 &0.768&0.534&0.692 \\
\hline
\end{tabular} \vspace{-3mm}
\end{table}\\
\subsection{UA-Backprop for A Deterministic NN} Our method can be applied to Ensemble-1 where the uncertainty is calculated by the entropy, i.e., the aleatoric uncertainty. However, the results shown in Table \ref{tab:single} are not good due to inadequate uncertainty quantification (UQ). By using more advanced single-network UQ methods, i.e. Laplacian approximation (LA)\cite{MacKay1992}, our UA method can yield improved results. Note that LA can also provide parameter samples from the posterior distribution, which can be directly used for UA-Backprop. Further studies on effectively performing UA on deterministic models can be our future direction. We will also concentrate on developing an end-to-end training approach that produces the attribution maps for a single network during training iterations and integrates the knowledge of UA for further model enhancement.

\subsection{Compare to Random Map}
In this section, we compare our proposed method with the random map to better illustrate the effectiveness of the proposed method. The random map is generated by sampling each element from the uniform distribution $U[0,1]$. To this end, the blurring test results are presented in Table \ref{tab:random} to compare the performance of the proposed method against the random maps. The experimental results show that the random maps fail to reduce the uncertainty during the blurring test.
\begin{table}[h]\small
	\caption{MURR and AUC-URR (AUC) of the blurring test to compare our proposed method with randomly generated maps. The number of blurring pixels is $2\%$ of the total pixels. The studies are conducted on MNIST and SVHN datasets.}
	\label{tab:random}
	\centering
\begin{tabular}{|l|cccc|l|}
\hline
	\multirow{3}{*}{Method} & \multicolumn{4}{|c|}{Dataset} \\
	\cline{2-5}
	& \multicolumn{2}{c}{MNIST (2\%)} & \multicolumn{2}{c|}{SVHN (2\%)} \\
 &MURR &AUC &MURR &AUC \\
\hline  
\multirow{1}{*}{Ours} &\textbf{0.648}  &\textbf{0.667} &\textbf{0.625}  &\textbf{0.526}   \\
\multirow{1}{*}{Random} &0.023&0.987&0.011&0.992 \\ 
\hline
\end{tabular}
\end{table}
\subsection{Compare to UA-Backprop without Normalization}
The normalization steps are required to achieve the completeness property. Nevertheless, an ablation study shows that with normalization, the MURR (2\% / 5\%) is \textbf{0.648}/\textbf{0.850} for MNIST and \textbf{0.629}/\textbf{0.848} for C10; without normalization, it can only achieve 0.471/0.797 for MNIST and 0.518/0.727 for C10.\\
\section{Additional Examples}
\begin{figure*}[t]
    \centering
    \includegraphics[width=1.0\linewidth]{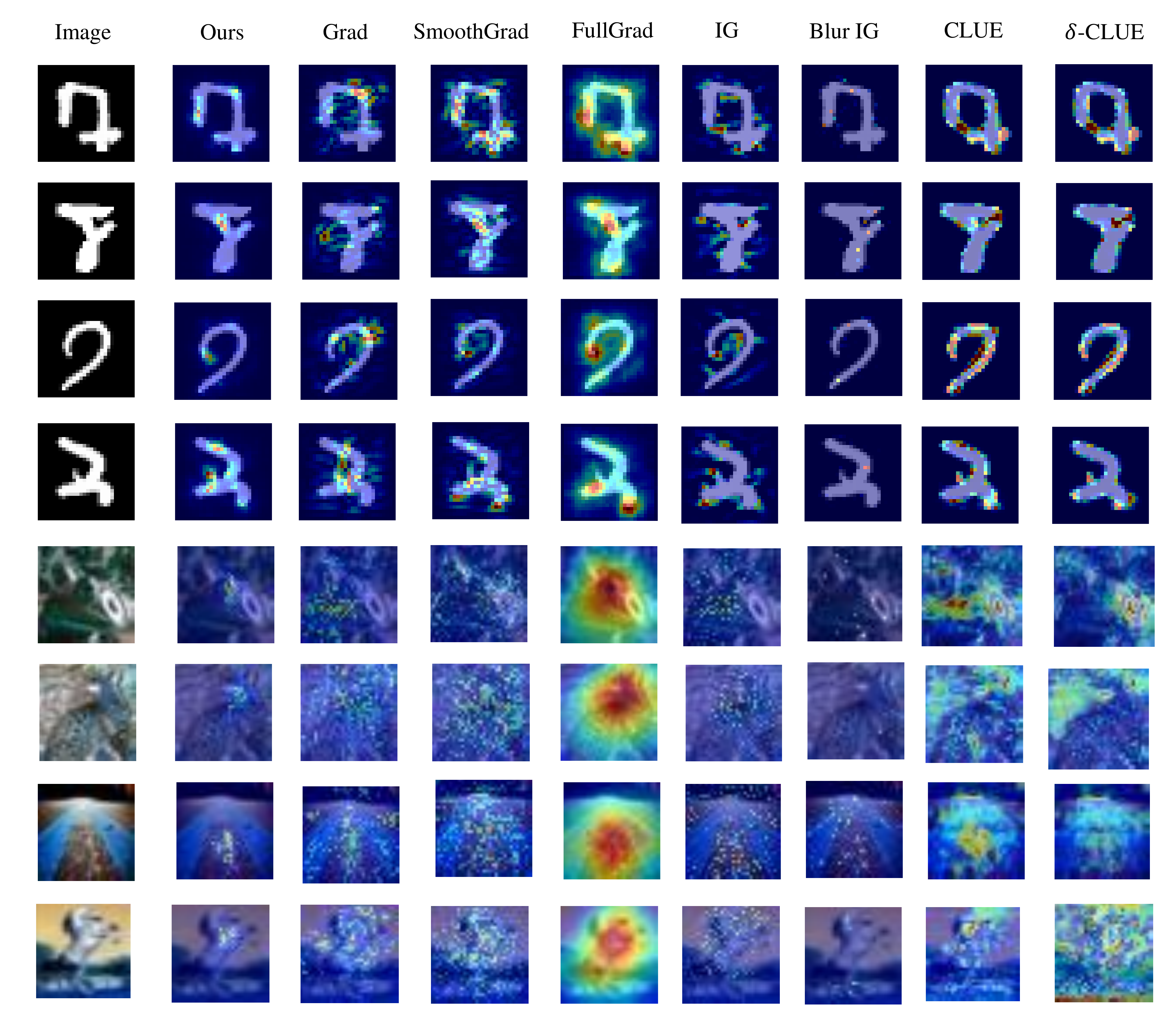}
    \caption{Additional examples of the uncertainty attribution maps for various methods across multiple datasets. 
    \label{fig:additional}}
\end{figure*}

Figure \ref{fig:additional} displays supplementary instances of the uncertainty attribution maps generated by various methods across multiple datasets. Our proposed method offers a more understandable and clear visualization of the generated maps compared to the vanilla application of existing CA methods. The latter often yields ambiguous explanations because of the presence of noisy gradients. In contrast, our approach provides a decomposition of pixel-wise contributions that efficiently explains the uncertainty while offering better regional illustrations that could be comprehended by individuals without expertise in the field.

\end{document}